\documentclass[]{interact}

\usepackage{epstopdf}
\usepackage{caption}
\usepackage{subcaption}

\usepackage{natbib}
\bibpunct[, ]{(}{)}{;}{a}{}{,}
\renewcommand\bibfont{\fontsize{10}{12}\selectfont}

\theoremstyle{plain}

\theoremstyle{definition}

\theoremstyle{remark}

\usepackage{graphicx} 
\usepackage{blindtext} 

\usepackage{algorithm}
\usepackage{algorithmic}

\usepackage{comment}
 
\newcommand{\todo}[1]{}

\newcommand{\change}[1]{}
\newcommand{\question}[1]{}

\RequirePackage{amsthm}
\usepackage{hyperref}
\usepackage{enumitem}
\usepackage{bbm}
\usepackage{color}
\usepackage{xcolor}
\usepackage{float}
\usepackage{chngcntr}
\usepackage{mathtools}
\usepackage{tikz}
\usepackage{tcolorbox}
\usepackage{cleveref}
\usepackage{cleveref}
\usepackage{booktabs}
\usepackage{multirow}
\usepackage{booktabs}
\usepackage{natbib}
\usepackage{mathptmx}
\usepackage{amsfonts}
\usepackage{amsmath}
\usepackage{amssymb}
\usepackage{amsthm}
\usepackage{pdflscape}

\usetikzlibrary{tikzmark, decorations.pathreplacing, calc, positioning, bayesnet}

\crefname{assumption}{assumption}{Assumptions}

\begin{document}


\title{Sparse Variational Contaminated Noise Gaussian Process Regression with Applications in Geomagnetic Perturbations Forecasting}

\author{
    \name{Daniel Iong\textsuperscript{a}, Matt McAnear\textsuperscript{a}, Yuezhou Qu\textsuperscript{a}, Shasha Zou\textsuperscript{b}, Gabor Toth\textsuperscript{b}, and Yang Chen\textsuperscript{a,c}\thanks{Contact: All correspondence can be sent to ychenang@umich.edu.}}
    \affil{
    \textsuperscript{a}Department of Statistics, University of Michigan; 
    \textsuperscript{b}Department of Climate and Space Sciences and Engineering, University of Michigan;
    \textsuperscript{c}Michigan Institute for Data Science, University of Michigan.
    }
}

\maketitle

\begin{abstract}
    Gaussian Processes (GP) have become popular machine learning methods for kernel based learning on datasets with complicated covariance structures. In this paper, we present a novel extension to the GP framework using a contaminated normal likelihood function to better account for 
    heteroscedastic variance and outlier noise.  We propose a scalable inference algorithm based on the Sparse Variational Gaussian Process (SVGP) method for fitting sparse Gaussian process regression models with contaminated normal noise on large datasets. We examine an application to geomagnetic ground perturbations, where the state-of-art prediction model is based on neural networks. We show that our approach yields shorter predictions intervals for similar coverage and accuracy when compared to an artificial dense neural network baseline.
\end{abstract}


\section{Introduction}
\label{sec:intro}

Gaussian process regression (GPR) is a popular nonparametric regression method
due to its ability to quantify uncertainty through the posterior predictive
distribution. GPR models can also incorporate prior knowledge through selecting an
appropriate kernel function. GPR commonly assumes a homoscedastic Gaussian distribution for observation
noise because this yields an analytical form for the posterior predictive prediction. However, Bayesian
inference based on Gaussian noise distributions is known to be sensitive to outliers
which are defined as observations that strongly deviate from model assumptions.

In regression, outliers can arise from relevant inputs being absent from
the model, measurement error, and other unknown sources. These outliers are associated with unconsidered sources of variation that affect the target variable 
sporadically. In this case, the observation model is unable
to distinguish between random noise and systematic effects not captured by the model.
In the context of GPR under Gaussian noise, outliers can heavily influence the posterior
predictive distribution, resulting in a biased estimate of the mean function and overly
confident prediction intervals.
Therefore, robust observation models are desired in the presence of potential outliers.

In this context, we consider geomagnetic perturbation measurements from
various ground magnetometer stations around the globe from 2010 to 2015 that we obtained
from SuperMAG \citep{gjerloev2012supermag}. This data provides a proxy for measuring geomagnetically induced currents, which could potentially drive catastrophic
disruptions to critical infrastructure, such as power grids and oil pipelines \citep{schrijver2014assessing}.
Therefore, it is imperative that we are able to obtain accurate predictions and
predictive uncertainty estimates for these quantities.

\begin{figure}
    \centering
    {\includegraphics[width=\columnwidth]{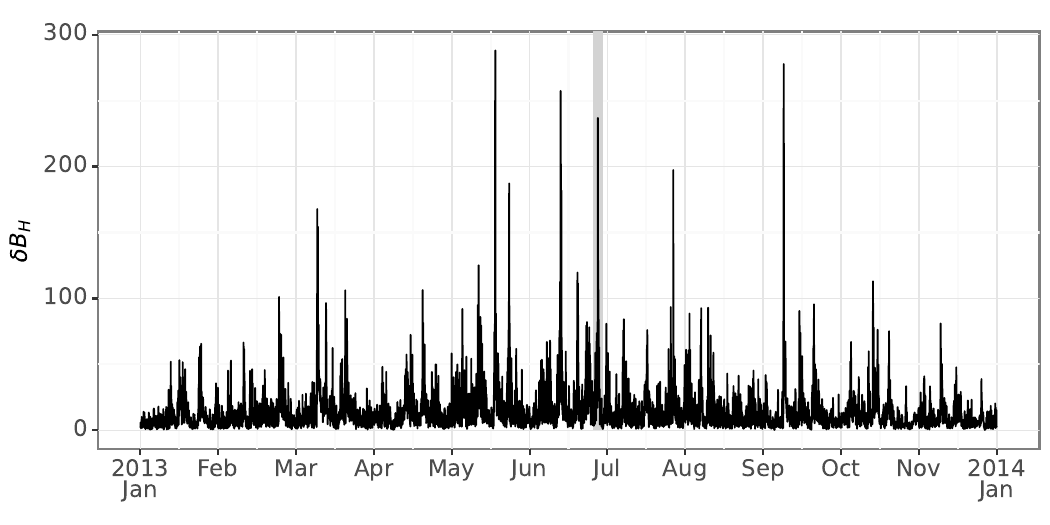}}
    \caption{Raw $\delta B_H$ values for the Ottawa magnetometer.}
    \label{fig:ott-values}
\end{figure}

Our goal is to predict the maximum value of the horizontal magnetic perturbation in the north-south direction,
denoted by $\delta B_{H}$, over twenty minute intervals across twelve test stations. 
This data is characterized by occasional large spikes that occur during geomagnetic
storms followed by long periods of relatively low variance. In this context, the influence of outliers is especially pronounced.  We plot $\delta B_{H}$ in
2013 in \cref{fig:ott-values} as an illustrative example. We use
solar wind and interplanetary magnetic field measurements obtained from NASA's OMNIWeb \citep{omni} as drivers of $\delta B$ (\cref{tab:features}).


\begin{table}
    \centering
    \begin{tabular}{ccc}
        \toprule
        Feature & Description & Units\\
        \midrule
        $\text{bx}_{gse}$ & X component of magnetic vector in GSE coordinates&  nT\\
        $\text{by}_{gse}$ & Y component of magnetic vector in GSE coordinates & nT\\
        $\text{bz}_{gse}$ & Z component of magnetic vector in GSE coordinates & nT\\
        v & Solar Wind Speed & km/s \\
        density & Proton Density & $N/cm^3$\\
        temperature & Proton Temperature & K \\
        pressure & Flow Pressure & nPa\\
        e & Electric Field & mV/m\\
    \end{tabular}
    \caption{Features used in prediction of $\delta$ B.}
    \label{tab:features}
\end{table}


To better account for the outliers seen in $\delta B_H$ prediction, we consider a GPR model with contaminated normal (CN) noise to account
for outliers. This stands in contrast to other ML methods, particularly artificial neural networks (ANNs). Often, the estimates from ANN models 
result in poorly calibrated coverage intervals that are too wide for a given confidence level, especially during periods of solar storms when outliers become 
relatively routine. To combat this, we developed a contaminated-normal distribution for Gaussiaan Processes that accounts for changing variance over storm periods.

The CN distribution is a
special case of a Gaussian mixture distribution with two components
\citep{gleason1993understanding}. This distribution models outliers explicitly by
assigning them to a mixture component with much larger variance. Therefore, the mixture
proportions can be interpreted as the outlier and non-outlier proportion.
Robust regression with mixture noise distributions is
not a new concept. \cite{box1968bayesian} first introduced a Bayesian linear regression model
with mixture noise as a Bayesian approach to handling outliers.
\cite{faul2001variational} follows up with a variational approximation for a similar
model. Although the idea of modeling noise with mixture distributions in regression
dates back to the 1960s, it is still being actively explored as a means to robustify
modern statistical and machine learning methods.
\cite{ferrari2018deepgum} proposed DeepGUM, a deep regression model with Gaussian-uniform
mixture noise for performing various computer vision tasks in the
presence of outliers. \cite{xu2019robust} developed the Mixture-of-Gaussian Lasso
technique which models noise as a Gaussian mixture distribution in a sparse linear
regression model. \cite{sadeghian2022robust} introduced a probabilistic principal
component regression model with switching Gaussian mixture noise for industrial process modeling.

GPR with mixture noise was first proposed by \cite{kuss2006gaussian} who developed an
expectation propagation (EP) algorithm and Markov chain Monte-Carlo (MCMC) sampling scheme for
inference. \cite{kuss2006gaussian} also developed inference algorithms for GPR models
with Student-t and Laplace noise. Their experiments showed that the robust GPR models
outperform the GPR model with Gaussian noise when outliers are present in the training
data. However, predictive performance among the robust GPR models were similar for the
simulated and real-world datasets they considered.
\cite{daemi2019gaussian} also considered a GPR model with mixture noise but used an
EM algorithm for inference. \cite{naish2007robust} proposed the twinned GP model which
also assumes a mixture noise distribution. However, they model the outlier proportion
with another GP, effectively making their noise model heteroscedastic which is out
of the scope of our proposed work. They showed that the twinned GP model is suitable in
cases where outliers can be clustered.
A major issue with the inference algorithms developed by \cite{kuss2006gaussian},
\cite{daemi2019gaussian}, and \cite{naish2007robust} is that they scale poorly with
number of observations as they perform inference on exact GPs which require inverting
a full kernel matrix.

In addition to GPR models with mixture noise, several other robust GPR methods have also
been proposed.
\cite{jylanki2011robust} provided a robust expectation propagation (EP) algorithm for
GPR with student-t noise. \cite{li2021robust} proposed iterative trimming for
robust GPR. The main idea of their method is to iteratively trim a proportion of the
observations with the largest absolute residuals so that they are not as influential to
the resulting model fit. However, this may not be appropriate in cases where outliers are
due to systematic effects not captured by the model but are still of interest to the analysis.
\cite{algikar2023robust} recently proposed using a Huber likelihood
for robust GPR. This method employs weights based on projection statistics to scale
residuals and bound the influence of outliers on the latent function estimate.
\cite{gu2018robust} discusses methods for robustly estimating the covariance parameters in a GP model with Gaussian noise. Unfortunately, all of these methods also perform exact inference. \cite{altamirano2023robust} proposes a robust and conjugate Gaussian process that also performs exact inference but can be plugged into scalable GP approximations. Their method involves doing generalized Bayesian inference with a robust loss function whereas our proposed method involves an explicit robust noise assumption.  

Many approximation methods for scaling up GPR to accomodate massive datasets have been proposed \citep{liu2019when}.
These approximation methods can be categorized as either global or local approximations.
Global approximations approximate the kernel matrix, and as a result the latent function posterior, through global distillation. This
can be achieved by computing the kernel matrix on only a subset of the training
data, sparsifying the kernel matrix, or constructing a sparse approximation of the latent
function posterior using a small number of inducing (or pseudo) points.
Local approximations take a divide and
conquer approach by focusing on local subsets of training data. While global
approximations are better at capturing global patterns, they often filter out local
patterns due to the use of inducing points. On the other hand, local approximations are
better at capturing non-stationary features but risk local over-fitting.
\cite{liu2019when} provides a comprehensive review of both global and local approximation methods for GPR.

Sparse approximations have perhaps received the most attention in the GP community.
Early work in this area involved approximating the GP prior with inducing
points and then optimizing the marginal likelihood of the approximate model \citep{quinonero-candela2005unifying}.
This includes the deterministic training conditional (DTC), fully
independent training conditional (FITC), and partially independent training conditional
(PITC) approximations which only differ in how they specify the dependency between
the latent function and inducing points
\citep{seeger2003fast,snelson2005sparse,snelson2007local}.
\cite{lazaro-gredilla2009inter-domain} further extended these approximated GP priors to be more
flexible by placing the inducing points in a different domain via integral transforms.
However, these approaches effectively alter the original model assumptions and may be
prone to overfitting. This motivated the development of approximate inference approaches
that make all the necessary approximations at inference time.

\cite{titsias2009variational} first introduced the sparse variational GP (SVGP)
method. A subtle but important distinction between this approach and earlier approaches
is that the inducing points are now considered variational parameters and are decoupled
from the original model. \cite{titsias2009variational} jointly estimates the inducing
points and model hyperparameters by maximizing a lower bound to the exact marginal
likelihood as opposed to maximizing an approximate marginal likelihood.
\cite{hensman2013gaussian} extended this idea by reformulating the lower bound to enable
stochastic optimization. \cite{hensman2015scalable} further modified this lower bound to
accomodate non-Gaussian likelihoods and applied it to GP classification. Sparse
variational methods are discussed in more detail in \cref{sec:sparse-vari-gp}.
In this paper, we propose a scalable inference algorithm based on the SVGP method for
fitting sparse GPR models with contaminated normal (CN) noise on large datasets.

The rest of the paper is organized as follows.
\Cref{sec:background} provides background on Gaussian process regression and sparse variational GPs. We then discuss our proposed model in \cref{sec:model} followed
by a corresponding inference algorithm in \cref{sec:inference}.
In \cref{sec:parameter-estimation}, we perform a simulation
study to show the efficacy of our proposed inference algorithm. In
\cref{sec:comp-to-sparse}, we compare sparse GPR models with different noise
distributions (CN, Gaussian, Student-t, Laplace) trained on simulated datasets.
In section \Cref{sec:applications}, we train GPR models with variable noise distributions on flight
delays and ground magnetic perturbations data and compare their predictive performance.
We conclude the paper with a brief summary and discussion on potential extensions in
\cref{sec:conc}. For all simulation studies and real-world data applications, we use the
GPyTorch Python package to train GPR models \citep{gardner2018gpytorch}.

\section{Background}
\label{sec:background}

In this section, we provide a primer on Gaussian process regression and review sparse variational GPs. 

\subsection{Gaussian process regression}
\label{sec:gauss-proc-regr}

A Gaussian process (GP) is mathematically defined as a collection of random variables
$\{f(\textbf{x}) | \textbf{x} \in \mathcal{X}\}$, for some index set $\mathcal{X}$, for
which any finite subset follows a joint Gaussian distribution. A GP is completely
specified by its mean and covariance (or kernel) function defined by
\begin{align*}
  m(\textbf{x}) &= E[f(\textbf{x})], \\
  k(\textbf{x}, \textbf{x}') &= E[(f(\textbf{x}) - m(\textbf{x}))(f(\textbf{x}') - m(\textbf{x}'))].
\end{align*}

GPs are commonly used as priors over real-valued functions. For the remainder
of this paper, we will assume that $\mathcal{X} = \mathbb{R}^d$ and $m(\textbf{x}) = 0$.
Suppose we have a training dataset consisting of inputs $\mathbf{X} = \{\textbf{x}_i\}_{i=1}^n$ and
outputs $\mathbf{y} = \{y_i\}_{i=1}^n$.
GP regression assumes that the outputs are noisy realizations of a latent function $f$
evaluated at the inputs, i.e. $y_i = f(\textbf{x}_i) + \epsilon_i$, where $\epsilon_i$
is a noise term. Furthermore, we assume that $\mathbf{f} = \{f(\mathbf{x}_{i})\}_{i=1}^{n}$ follows a Gaussian process prior, i.e.
\begin{equation}
  p(\mathbf{f} | \vartheta) = N(\mathbf{f} | \mathbf{0}, \mathbf{K}_{nn}),
  \label{eq:gp-prior}
\end{equation}
where
$\mathbf{K}_{nn} = \{k(\textbf{x}_{i}, \textbf{x}_{j}; \vartheta)\}_{1 \leq i,j \leq n} \in \mathbb{R}^{n \times n}$
is a covariance matrix with kernel hyperparameters $\vartheta$. For the rest of this
paper, we will denote the Gaussian density with mean $\mu$ and variance $\sigma^{2}$ as
$N(\cdot | \mu, \sigma^{2})$.
We will drop the dependence
on $\vartheta$ in our notation when it does not need to be emphasized. Combined
with a likelihood function $p(\textbf{y} | \textbf{f})$, or equivalently a distribution
for the noise term,
the joint distribution between outputs and latent
function values fully specifies the GPR model and takes the form
\begin{equation}
  p(\mathbf{y}, \mathbf{f}) = p(\mathbf{y} | \mathbf{f}) p(\mathbf{f}).
  \label{eq:joint-distr}
\end{equation}
Note that this distribution and subsequent distributions will depend on inputs $\textbf{X}$ and
potential model/kernel hyperparameters but we will omit them in our notation whenever
this dependence does not need to be emphasized.
The likelihood is usually assumed to follow a Gaussian distribution with homoskedastic noise centered at $\textbf{f}$, i.e.
\begin{equation}
  p(\textbf{y} | \textbf{f}) = N(\mathbf{y} | \textbf{f}, \sigma^2 \mathbf{I}_n) = \prod_{i=1}^{n} p(y_{i} | f(\mathbf{x}_{i}), \sigma^{2}),
  \label{eq:gaussian-noise}
\end{equation}
where $\mathbf{I}_{n}$ denotes the $n \times n$ identity matrix.
In \cref{sec:meth}, we will replace this with a CN distribution to account for outliers or extreme observations.
From the joint probability model specified by \cref{eq:joint-distr}, we can derive
several distributions of interest, namely the marginal likelihood and posterior
predictive distribution.
These distributions can be derived in closed form if we make
the Gaussian noise assumption in \cref{eq:gaussian-noise}.
Marginalizing out $\textbf{f}$ in \cref{eq:joint-distr}, we get that the marginal
likelihood is given by
\begin{equation}
    p(\textbf{y}) = N(\textbf{y} | \mathbf{0}, \mathbf{K}_{nn} + \sigma^2 \mathbf{I}_n).
\end{equation}
Maximum likelihood estimates of hyperparameters can be obtained by maximizing the
marginal likelihood with respect to the hyperparameters using gradient-based optimization.
The posterior process is also a GP with the following posterior mean and
covariance functions:
\begin{align*}
  m_{\mathbf{y}}(\mathbf{x}) &= \mathbf{k}_{\mathbf{x}n}' \tilde{\mathbf{K}}_{nn}^{-1} \mathbf{y}, \\
  k_{\mathbf{y}}(\mathbf{x}, \mathbf{x}') &= k(\mathbf{x}, \mathbf{x}') - \mathbf{k}_{\mathbf{x}n}' \tilde{\mathbf{K}}_{nn}^{-1} \mathbf{k}_{\mathbf{x}'n},
\end{align*}
where
$\mathbf{k}_{\mathbf{x}n} = \big(k(\mathbf{x}, \mathbf{x}_{1}), \dots, k(\mathbf{x}, \mathbf{x}_{n}) \big)'$
and $\tilde{\mathbf{K}}_{nn} = \mathbf{K}_{nn} + \sigma^{2} \mathbf{I}_{n}$. In other words,
$p \big(f(\mathbf{x}) | \mathbf{y} \big) = N \big(f(\mathbf{x}) \big| m_{\mathbf{y}}(\mathbf{x}), k_{\mathbf{y}}(\mathbf{x}, \mathbf{x}) \big)$
for any $\mathbf{x} \in \mathbb{R}^{d}$.
After obtaining hyperparameter
estimates, we can make predictions at a test input $\mathbf{x}^{*}$ using the posterior
predictive distribution:
\begin{align}
  p(y^{*} | \mathbf{y}, \mathbf{x}^{*}) &= \int p(y^{*} | f^{*}) p(f^{*} | \mathbf{y}) df^{*} \nonumber \\
  &= N(y^{*} | m_{\mathbf{y}}(\mathbf{x}^{*}), \sigma^{2} + k_{\mathbf{y}}(\mathbf{x}^{*}, \mathbf{x}^{*})).
    \label{eq:gaussian-post-pred}
\end{align}
A more comprehensive review of Gaussian process regression can be found in \cite{rasmussen2005gaussian}.
Computing \cref{eq:gaussian-post-pred} involves inverting
the $n \times n$ matrix $\tilde{\mathbf{K}}_{nn}$ which requires $O(n^{3})$ computations. This
restricts the use of exact GPR models to datasets with up to a few thousand observations.
Sparse approximations can be used to make GPR more scalable.
Sparse GP methods approximate the latent function posterior by conditioning on a small number
$m << n$ of inducing points that act as a representative proxy for the observed
outputs. These inducing points can either be chosen as a subset of the training data or
optimized over. In the remainder of this section, we will discuss popular methods
for jointly estimating the inducing points and hyperparameters.

%
%
%
%
Let $\mathbf{Z} = \{\mathbf{z}_{i}\}_{i=1}^{m}$ denote a set of inducing inputs, which
may not be identical to the original inputs, and define the corresponding inducing points as
$\mathbf{u} = \{f(\mathbf{z}_{i})\}_{i=1}^{m}$. It follows that
$p(\textbf{u}) = N(\textbf{u} | \mathbf{0}, \mathbf{K}_{mm})$, where $\mathbf{K}_{mm}$ is defined analogously to $\mathbf{K}_{nn}$.
The joint distribution between $\mathbf{f}$ and $\mathbf{u}$ is given by $p(\textbf{f}, \textbf{u}) = p(\textbf{f} | \textbf{u}) p(\textbf{u}),$
where
\begin{equation}
  p(\textbf{f} | \textbf{u}) = N(\textbf{f} | K_{nm} \mathbf{K}_{mm}^{-1} \textbf{u}, \mathbf{K}_{nn} - \mathbf{Q}_{nn}),
  \label{eq:p-f-u}
\end{equation}
with $\mathbf{Q}_{nn} = \mathbf{K}_{nm} \mathbf{K}_{mm}^{-1} \mathbf{K}_{nm}'$ and $\mathbf{K}_{nm} = \{ k(\textbf{x}_i, \textbf{z}_j) \} \in \mathbb{R}^{n \times m}$.
The original joint probability model in \cref{eq:joint-distr} can then be augmented with
the inducing points to form the model
\begin{equation}
  p(\mathbf{y}, \mathbf{f}, \mathbf{u}) = p(\mathbf{y} | \mathbf{f}) p(\mathbf{f} | \mathbf{u}) p(\mathbf{u}).
  \label{eq:augmented-model}
\end{equation}
Note that integrating out $\mathbf{u}$ returns us to the original model so they are
equivalent in terms of performing inference. However, inference with this augmented
model still requires inverting $n \times n$ matrices.
Several methods have been proposed to approximate the distribution in \cref{eq:p-f-u}, thereby modifying the GP prior
and likelihood \citep{quinonero-candela2005unifying}. The fully
independent training conditional (FITC) approximation removes the conditional
dependencies between different elements in $\mathbf{f}$, i.e.
\begin{equation}
  q(\mathbf{f} | \mathbf{u}) = N \big(\mathbf{f} | \mathbf{K}_{nm} \mathbf{K}_{mm}^{-1} \mathbf{u}, \text{diag}(\mathbf{K}_{nn} - \mathbf{Q}_{nn}) \big),
  \label{eq:fitc-f-u}
\end{equation}
where $\text{diag}(\mathbf{A})$ denotes a diagonal matrix formed with the diagonal elements of
$\mathbf{A}$. This results in the following modified marginal likelihood:
\begin{equation*}
  \tilde{p}(\mathbf{y} | \mathbf{f}) = N(\mathbf{y} | 0, \sigma^{2} \mathbf{I}_{n} + \tilde{\mathbf{Q}}_{nn}),
\end{equation*}
where $\tilde{\mathbf{Q}}_{nn} = \mathbf{Q}_{nn} + \text{diag}(\mathbf{K}_{nn} - \mathbf{Q}_{nn})$ is an approximation to
the true covariance $\mathbf{K}_{nn}$.
Similarly, the deterministic training conditional (DTC) and partially
independent training conditional (PITC) methods approximate $\mathbf{K}_{nn}$ with
$\mathbf{Q}_{nn}$ and a block diagonalization of $\mathbf{K}_{nn} - \mathbf{Q}_{nn}$, respectively.
The inducing points and hyperparameters can then be jointly estimated by maximizing the
marginal log-likelihood.
With these approximations, the cost of inference and prediction is reduced from
$O(n^{3})$ to $O(nm^{2})$. However, these methods are philosophically troubling as they
entangle assumptions about the data embedded in the original likelihood with the
approximations required to perform inference. Furthermore, $m$ new model parameters are added
which increases the risk of overfitting. The sparse variational GP (SVGP) method, first
introduced by \cite{titsias2009variational}, takes a different approach by approximating
the exact posterior GP with variational inference. Before discussing this approach, we
take a detour to give the reader a primer on variational inference.


\subsection{Sparse variational GPs (SVGP)}
\label{sec:sparse-vari-gp}

In contrast to the DTC, FITC, and PITC approximations, the SVGP method performs
inference with the exact augmented model in \cref{eq:augmented-model} and approximates
the posterior distribution
$p(\mathbf{f}, \mathbf{u} | \mathbf{y}) = p(\mathbf{f} | \mathbf{u}, \mathbf{y}) p(\mathbf{u} | \mathbf{y})$
using variational inference. \Cref{sec:vari-infer} provides a brief introduction to variational inference. In other words, we want to solve the following optimization
problem:
\begin{equation*}
  q^{*}(\mathbf{f}, \mathbf{u}) = \underset{q}{\text{arg min }} \text{KL}\big[ q(\mathbf{f}, \mathbf{u}) || p(\mathbf{f}, \mathbf{u} | \mathbf{y}) \big].
\end{equation*}
To ensure efficient computation, the approximate posterior is assumed to factorize as
\begin{equation}
  q(\mathbf{f}, \mathbf{u}) = p(\mathbf{f} | \mathbf{u}) q(\mathbf{u}),
  \label{eq:opt-q-form}
\end{equation}
where $p(\mathbf{f} | \mathbf{u})$ is given in \cref{eq:p-f-u}; and $q(\mathbf{u})$ is a variational distribution for $\mathbf{u}$.
To see why, let's compute the ELBO from \cref{eq:elbo-def}:
\begin{equation*}
  \text{ELBO} = E_{q(\mathbf{f}, \mathbf{u})} \Big[ \log \frac{p(\mathbf{y} | \mathbf{f}) p(\mathbf{f} | \mathbf{u}) p(\mathbf{u})}{q(\mathbf{f}, \mathbf{u})} \Big].
\end{equation*}
If we assume $q(\mathbf{f}, \mathbf{u})$ takes the form in \cref{eq:opt-q-form}, then
the $p(\mathbf{f} | \mathbf{u})$ terms in the log cancel out and the ELBO can be computed as
\begin{align}
  \text{ELBO} &=  E_{p(\mathbf{f} | \mathbf{u}) q(\mathbf{u})} \Big[ \log \frac{p(\mathbf{y} | \mathbf{f}) p(\mathbf{u})}{q(\mathbf{u})} \Big] \label{eq:elbo-svgp} \\
              &= E_{q(\mathbf{u})} \Big[ E_{p(\mathbf{f} | \mathbf{u})}[\log p(\mathbf{y} | \mathbf{f})] + \log \frac{p(\mathbf{u})}{q(\mathbf{u})} \Big] \nonumber \\
              &= E_{q(\mathbf{f})} [\log p(\mathbf{y} | \mathbf{f})] - \text{KL} [q(\mathbf{u}) || p(\mathbf{u})] \nonumber \\
  &= \sum_{i=1}^{n} E_{q(f_{i})} [\log p(y_{i} | f_{i})] - \text{KL} [q(\mathbf{u}) || p(\mathbf{u})] = \mathcal{L}_{\text{svgp}}, \label{eq:elbo-factorized}
\end{align}
where $f_{i} = f(\textbf{x}_{i})$ and $q(\mathbf{f}) = \int p(\mathbf{f} | \mathbf{u}) p(\mathbf{u}) d\mathbf{u}$. This
can be computed with $O(nm^{2})$ computations. \cite{titsias2009variational} showed that
under a Gaussian likelihood, this ELBO can be maximized without explicitly computing the
optimal $q(\mathbf{u})$ by maximizing the lower bound:
\begin{equation*}
  \log N(\mathbf{y} | 0, \sigma^{2} \mathbf{I} + \mathbf{Q}_{nn}) - \frac{1}{2 \sigma^{2}} \text{Tr}(\tilde{\mathbf{K}}_{nn}),
\end{equation*}
where $\tilde{\mathbf{K}}_{nn} = \mathbf{K}_{nn} - \mathbf{Q}_{nn}$; and $\text{Tr}(\mathbf{A})$ denotes the trace of matrix $\mathbf{A}$. The derivation for this lower bound is given in \cref{sec:elbo-deriv1}.
In order to maximize the ELBO using stochastic variational inference,
\cite{hensman2013gaussian} proposed maintaining an explicit variational distribution
given by
\begin{equation}
  q(\mathbf{u}) = N(\mathbf{u} | \mathbf{m}, \mathbf{S}).
  \label{eq:q-u}
\end{equation}
This yields the following form for $q(\mathbf{f})$:
\begin{equation}
  q(\mathbf{f}) = N(\mathbf{f} | \mathbf{A}\mathbf{m}, \mathbf{K}_{nn} + \mathbf{A}(\mathbf{S} - \mathbf{K}_{mm}) \mathbf{A}'),
  \label{eq:var-post-f}
\end{equation}
where $\mathbf{A} = \mathbf{K}_{nm} \mathbf{K}_{mm}^{-1}$. Under a Gaussian likelihood, the ELBO becomes
\begin{equation*}
  \sum_{i=1}^{n} \Big[ \log N(y_{i} | \mathbf{k}_{i}' \mathbf{K}_{mm}^{-1} \mathbf{m}, \sigma^{2}) - \frac{1}{2} \sigma^{-2} \tilde{\mathbf{k}}_{ii} - \frac{1}{2} \text{Tr}(\mathbf{S} \boldsymbol{\Lambda}_{i}) \Big] - \text{KL}[q(\mathbf{u}) || p(\mathbf{u})],
\end{equation*}
where $\mathbf{k}_{i}$ is the $i$th column of $\mathbf{K}_{mn}$; $\tilde{\mathbf{k}}_{ii}$ is the $i$th diagonal
element of $\tilde{\mathbf{K}}_{nn}$; and
$\boldsymbol{\Lambda}_{i} = \sigma^{-2} \mathbf{K}_{mm}^{-1} \mathbf{k}_{i} \mathbf{k}_{i}' \mathbf{K}_{mm}^{-1}$ (Section 3.1 of \cite{hensman2013gaussian}).
Furthermore, the approximate predictive distribution at a new input $\mathbf{x}^{*}$ is
given by
\begin{align}
  q_{\text{pred}}(y^{*} | \mathbf{x}^{*}) &= \int p(y^{*} | \mathbf{f}^{*}) q(\mathbf{f}^{*}) d\mathbf{f}^{*} \nonumber \\
  &= N(y^{*} | \mu_{f^{*}}, \sigma_{f^{*}}^{2} + \sigma^{2}), \label{eq:aprox-pred-gaussian}
\end{align}
where $\mu_{f^{*}} = k_{*}' \mathbf{K}_{mm}^{-1} \mathbf{m}$,
$\sigma_{f^{*}}^{2} = k(\mathbf{x}^{*}, \mathbf{x}^{*}) + \mathbf{k}_{*}' \mathbf{A} \mathbf{k}_{*}$, and
$\mathbf{k}_{*} = \{k(\mathbf{x}^{*}, \mathbf{x}_{i})\}_{i=1}^{n}$.
$\mathcal{L}_{\text{svgp}}$ can be computed for non-Gaussian likelihoods as long as the
expectation in \cref{eq:elbo-factorized} can be computed or approximated. Inference with
SVGP involves maximizing $\mathcal{L}_{\text{svgp}}$ with respect to variational,
likelihood, and kernel hyperparameters. In \cref{sec:inference}, we show how this method
can be extended to perform inference in GPR with CN noise.

\section{Methods}
\label{sec:meth}


\subsection{Model specification}
\label{sec:model}

We propose to replace the Gaussian noise assumption made in \cref{eq:gaussian-noise}
with the following contaminated normal (CN) noise assumption:
\begin{equation}
  p(y_i | f_i) = \pi N(y_i | f_i, \tau \sigma^2) + (1 - \pi) N(y_i | f_i, \sigma^2), \ \ i = 1, \dots, n,
  \label{eq:scn-noise}
\end{equation}
where $\sigma^{2} > 0$ is the noise variance for non-outlier observations;
$\tau > 1$ is an inflation parameter which represents the increased variance due to
outliers; and $0 < \pi < 1$ gives the proportion of outliers. In contrast to other
robust likelihoods such as the Laplace or Student-t likelihood, this likelihood
explicitly models the outliers by giving them an outsized variance compared to
non-outlier or extreme observations. This is motivated by the needs of modeling occasional extreme phenomena in geomagnetic disturbances as shown in Figure~\ref{fig:ott-values} in the Introduction.  


\subsection{Inference}
\label{sec:inference}

Let $\theta = (\pi, \tau, \sigma^{2})$ denote our model hyperparameters.
A naive approach to estimating $\theta$ is to directly plug \cref{eq:scn-noise} into $\mathcal{L}_{\text{svgp}}$ in \cref{eq:elbo-factorized} and
maximize it with respect to $\theta$ using gradient-based optimization. However, the
expectation term in \cref{eq:elbo-factorized} would not have a closed form and would need to be approximated.
In this section, we derive a modified ELBO based on \cref{eq:elbo-factorized} for our proposed model and introduce a stochastic generalized alternating maximization (SGAM) algorithm to maximize it in order to estimate $\theta$ and other hyperparameters. 
We begin by introducing a set of binary latent variables $\{\alpha_{i}\}_{i=1}^{n}$ that represent the
component assignments for each observation, i.e.
\begin{equation}
  p(\alpha_{i} = 1 | \theta) = \pi, \quad p(y_{i} | f_{i}, \alpha_{i}, \theta) = N(y_{i} | f_{i}, \tau^{\alpha_{i}} \sigma^{2}).
\end{equation}
Following the same logic in \cref{sec:vari-infer}, the log-likelihood function $\log p(y_{i} | f_{i}, \theta)$ can be decomposed as
\begin{equation*}
  \log p(y_{i} | f_{i}, \theta) = F(q(\alpha_{i} | f_{i}), \theta) + \text{KL}\big[q(\alpha_{i} | f_{i}) || p(\alpha_{i} | y_{i}, f_{i}, \theta)\big],
\end{equation*}
for any distribution $q(\alpha_{i} | f_{i})$, where
\begin{align*}
    F(q(\alpha_{i} | f_{i}), \theta) &= E_{q(\alpha_{i} | f_{i})} \Big[ \log \Big(\frac{p(y_{i}, \alpha_{i} | f_{i}, \theta)}{q(\alpha_{i} | f_{i})}  \Big)  \Big] \\
    &= E_{q(\alpha_{i} | f_{i})} [\log p(y_{i}, \alpha_{i} | f_{i}, \theta)] + H(q(\alpha_i | f_i));
\end{align*}
and $H(p(x)) = - \int p(x) \log p(x) dx$ is the entropy function. 
Since the KL divergence is
non-negative, $F(q(\alpha_{i} | f_{i}), \theta)$ provides a lower bound for the log-likelihood
function. By extension of \cref{eq:elbo-factorized}, this gives us the following modified ELBO:
\begin{align}
  \log p(\mathbf{y} | \Theta) &\geq \sum_{i=1}^{n} E_{q(f_{i}, \alpha_{i} | \varphi, \vartheta)} \big[ F(q(\alpha_{i} | f_{i}), \theta) \big] - \text{KL}[q(\mathbf{u}) || p(\mathbf{u}); \vartheta] \nonumber \\
                              &= \sum_{i=1}^{n} \Big\{ E_{q(f_{i}, \alpha_{i} | \varphi, \vartheta)} \big[ \log p(y_{i}, \alpha_{i} | f_{i}, \theta) \big] \\
                              & + E_{q(f_i | \varphi, \vartheta)} \big[ H(q(\alpha_{i} | f_{i})) \big] \Big\} - \text{KL}[q(\mathbf{u}) || p(\mathbf{u}); \vartheta] \nonumber \\
                              &= \mathcal{L}(q, \Theta), \label{eq:cn-elbo}
\end{align}
where $\Theta = (\theta, \vartheta)$ denotes the model and kernel hyperparameters and
$q(f_{i}, \alpha_{i} | \varphi, \vartheta) = q(\alpha_{i} | f_{i}) q(f_{i} | \varphi, \vartheta)$.
We assume $q(f_{i} | \varphi, \vartheta)$ is the Gaussian variational distribution given
in \cref{eq:var-post-f}. Although the true latent function posterior may have
multiple modes in this case, \cite{kuss2006gaussian} showed that a Laplace approximation for the
latent function posterior of an exact GP with mixture noise works reasonably well in the
presence of a few outliers. Since $q(\textbf{u})$ and $p(\textbf{u})$ are both Gaussian, the KL divergence term can be expressed as a sum of $n$ terms and therefore, $\mathcal{L}(q, \Theta)$ can be written as 
$\mathcal{L}(q, \Theta) = \sum_{i=1}^{n} \mathcal{L}_i(q, \Theta)$.  
This allows it to be maximized using stochastic optimization methods. For the rest of this section, we describe the SGAM algorithm for maximizing the modified ELBO (or equivalently, minimizing the negative modified ELBO). We summarize the algorithm in \cref{alg:sgam-pseudocode}.


%
%

Let $\Theta^{(0)} = (\theta^{(0)}, \vartheta^{(0)})$ denote the starting hyperparameter values. Furthermore, let
$q^{(0)}(f_{i}, \alpha_{i}) = q(\alpha_{i} | f_{i}) q(f_{i} | \varphi^{(0)}, \vartheta^{(0)})$ denote
the starting variational distribution with hyperparameters $\varphi^{(0)}$.
At each iteration $t \geq 1$, the SGAM algorithm alternates between a forward and backward step.
Let $I_t$ be a random subset uniformly sampled from $\{1, \dots, n\}$.
In the forward step, we update $q(f_i, \alpha_i)$, with $\Theta = \Theta^{(t-1)}$,
by first updating $q(\alpha_i | f_i)$ with
\begin{equation*}
  q^{(t)}(\alpha_{i} | f_{i}) = p(\alpha_{i} | f_{i}, y_{i}, \Theta^{(t-1)}).
\end{equation*}
This is equivalent to the E-step update in the standard EM algorithm where $f_i$ is held fixed. Furthermore, it maximizes $\mathcal{L}(q, \Theta^{(t-1)})$ with respect to $q(\alpha_i | f_i)$. Next, we obtain $\varphi^{(t)}$ by taking a stochastic gradient descent (SGD) step with respect to $\varphi$, i.e. 
\begin{equation}
    \varphi^{(t)} \leftarrow \varphi^{(t-1)} + \alpha_1^{(t)} \nabla_{\varphi} \mathcal{L}_{I_t}(\tilde{q}_{\varphi}^{(t)}, \Theta^{(t-1)}), \label{eq:varphi-sgd}
\end{equation}
where  $\tilde{q}_{\varphi}^{(t)}(f_i, \alpha_i) = q^{(t)}(\alpha_i | f_i) q(f_i | \varphi, \vartheta^{(t-1)})$;
$\alpha^{(t)}_{1}$ denotes the learning rate at iteration $t$; $\mathcal{L}_{I_t} = \sum_{i \in I_t} \mathcal{L}_i$; and $\nabla_{\varphi}$ denotes the gradient operator with respect to $\varphi$.
The resulting forward step update is given by
\begin{equation*}
  q^{(t)}(f_{i}, \alpha_{i}) = q(f_{i} | \varphi^{(t)}, \vartheta^{(t-1)}) p(\alpha_{i} | f_{i}, y_{i}, \Theta^{(t-1)}).
\end{equation*}

In the backward step, we update $\Theta$ by first maximizing 
\begin{align}
  \mathcal{L}(q^{(t)}, \Theta) = \rm{const.}+ \sum_{i=1}^{n} &- \frac{1}{2} \Big[ \log(\sigma^{2}) + \hat{\alpha}_{i}^{(t)} \log(\tau) + \nonumber \\
  &\sigma^{-2} \big(1 + (\tau^{-1} -1) \hat{\alpha}_{i}^{(t)} \big) D_{i}^{(t)} \Big]  \nonumber \\
  &+ \hat{\alpha}_{i}^{(t)} \log(\pi) + (1 - \hat{\alpha}_{i}^{(t)}) \log(1 - \pi) \label{eq:obj-M-step}
\end{align}
with respect to $\theta$,
where
\begin{align*}
  \hat{\alpha}_{i}^{(t)} &= \int p(\alpha_{i} = 1 | f_{i}, y_{i}, \Theta^{(t-1)}) q(f_{i} | \varphi^{(t)}, \vartheta^{(t-1)}) df_{i} \\
  &= \frac{\pi^{(t-1)} \Psi_{i}^{(t)} \big(\tau^{(t-1)} \sigma^{2(t-1)} \big)}{\pi^{(t-1)} \Psi_{i}^{(t)} \big(\tau^{(t-1)} \sigma^{2(t-1)} \big) + (1 - \pi^{(t-1)}) \Psi_{i}^{(t)} \big( \sigma^{2(t-1)} \big)},
\end{align*}
with $D_{i}^{(t)} = E_{q^{(t)}(f_{i})} \big[ (y_{i} - f_{i})^{2} \big] = (y_{i} - \mu_{f_{i}}^{(t)})^{2} + \sigma_{f_{i}}^{2(t)}$ and
$\Psi_{i}^{(t)}(x) = N(y_{i} | \mu_{f_{i}}^{(t)}, \sigma_{f_{i}}^{(t)} + x)$. 
It can be shown that the following closed-form updates maximize \cref{eq:obj-M-step}:
\begin{align}
  \pi^{(t)} &= \frac{N_{i}^{(t)}}{n}, \nonumber \\
  \sigma^{2(t)} &= \frac{1}{n} \sum_{i=1}^{n} \Big(1 + \big((\tau^{(t-1)})^{-1} - 1 \big) \hat{\alpha}_{i}^{(t)} \Big) D_{i}^{(t)}, \nonumber \\
  \tau^{(t)} &= \frac{1}{\sigma^{2(t)} N_{i}^{(t)}} \sum_{i=1}^{n} \hat{\alpha}_{i}^{(t)} D_{i}^{(t)}, \label{eq:tau-update}
\end{align}
where $N_{i}^{(t)} = \sum_{i=1}^{n} \hat{\alpha}_{i}^{(t)}$. 
Similar to the forward step, the second part of the backward step involves taking a SGD step with respect to $\vartheta$, i.e.
\begin{equation*}
    \vartheta^{(t)} \leftarrow \vartheta^{(t-1)} + \alpha_2^{(t)} \nabla_{\vartheta} \mathcal{L}_{I_t}(q^{(t)}, \tilde{\Theta}^{(t)}),
\end{equation*}
where $\tilde{\Theta}^{(t)} = (\theta^{(t)}, \vartheta)$; and $\alpha_2^{(t)}$ is the learning rate at iteration $t$, which is set to decay exponentially starting with an initial value of 0.1. To make this more computationally efficient, we can also take a SGD step with respect to $\theta$ to obtain $\theta^{(t)}$ instead of computing the closed-form updates in the first part of the backward step. 
Note that the $\tau^{(t)}$ update in
\cref{eq:tau-update} is not constrained to be greater than 1.
If the final $\tau$ estimate is less than 1, we can replace $\sigma^{2}$ with $\tau \sigma^{2}$, $\tau$ with $\tau^{-1}$, and $\pi$ with $1 - \pi$
so that $\tau$ and $\pi$ can still be interpreted as the variance inflation
parameter and outlier probability, respectively. The approximate posterior predictive distribution
at a new input $\mathbf{x}^{*}$ is given by
\begin{equation}
  q_{\text{pred}}(y^{*} | \mathbf{y}, \mathbf{x}^{*}) = \pi N(y^{*} | \mu_{f^{*}}, \sigma_{f^{*}}^{2} + \tau \sigma^{2}) + (1 - \pi) N(y^{*} | \mu_{f^{*}}, \sigma_{f^{*}}^{2} + \sigma^{2}),
  \label{eq:pred-distr-scn}
\end{equation}
where $\mu_{f^{*}}$ and $\sigma_{f^{*}}^{2}$ are the same as in \cref{eq:aprox-pred-gaussian}.

\begin{algorithm}
\caption{Stochastic Generalized Alternating Maximization (SGAM)}
\label{alg:sgam-pseudocode}
\begin{algorithmic}[1]
\STATE Initialize $\Theta^{(0)} = (\theta^{(0)}, \vartheta^{(0)})$
\STATE Initialize variational distribution $q^{(0)}(f_i, \alpha_i) = q(\alpha_i|f_i)q(f_i|\phi^{(0)}, \vartheta^{(0)})$
\FOR{$t = 1, 2, \ldots$}
    \STATE Sample a random subset $I_t$ uniformly from $\{1, \ldots, n\}$
    \STATE \textbf{Forward Step:}
    \FOR{each $i \in I_t$}
        \STATE Update $q^{(t)}(\alpha_i|f_i) = p(\alpha_i|f_i, y_i, \Theta^{(t-1)})$
    \ENDFOR
    \STATE Update $\phi^{(t)}$ using stochastic gradient descent:
    \[
    \phi^{(t)} \leftarrow \phi^{(t-1)} + \alpha_1^{(t)} \nabla_\phi L_{I_t}(q_\phi^{(t)}, \Theta^{(t-1)})
    \]
    \STATE Compute $q^{(t)}(f_i, \alpha_i) = q(f_i|\phi^{(t)}, \vartheta^{(t-1)}) p(\alpha_i|f_i, y_i, \Theta^{(t-1)})$
    
    \STATE \textbf{Backward Step:}
    \STATE Update $\Theta$ by maximizing $L(q^{(t)}, \Theta)$ with respect to $\theta$ and $\vartheta$:
    \[
    \Theta^{(t)} = \arg\max_\Theta L(q^{(t)}, \Theta)
    \]
    \STATE Alternatively, use stochastic gradient descent to update $\Theta$:
    \[
    \vartheta^{(t)} \leftarrow \vartheta^{(t-1)} + \alpha_2^{(t)} \nabla_\vartheta L_{I_t}(q^{(t)}, \Theta)
    \]
\ENDFOR
\end{algorithmic}
\end{algorithm}

\section{Simulation Studies}
\label{sec:simulation-studies}


In this section, we perform several simulation studies to evaluate the performance of
our proposed method and to compare it to similar methods. In the first simulation
study, we show that our proposed algorithm is able to recover ground truth parameter and
latent function values. In the second simulation study, we compare sparse variational GPR models
with Gaussian (\textbf{GPR-G}), contaminated normal (\textbf{GPR-CN}), Student-t
(\textbf{GPR-t}), and Laplace (\textbf{GPR-L}) distributed noise on simulated data
with varying degrees of outlierness. We show that GPR-CN outperforms the other
considered methods when there is a non-negligible proportion of outliers.  We evaluate the four
methods using the root mean squared error (RMSE), mean absolute error (MAE), and negative log
predictive density (NLPD) metrics. We use the RMSE and MAE to evaluate the
predictive mean functions. To evaluate the predictive distribution, we use the NLPD metric.
The RMSE between observed
values $\mathbf{y}$ and predicted values $\hat{\mathbf{y}}$ is defined as
\begin{equation*}
  \text{RMSE}(\mathbf{y}, \hat{\mathbf{y}}) = \sqrt{\frac{1}{n} \sum_{i=1}^{n} (y_{i} - \hat{y}_{i})^{2}}.
\end{equation*}
The MAE between $\mathbf{y}$ and $\hat{\mathbf{y}}$ is defined as
\begin{equation*}
  \text{MAE}(\mathbf{y}, \hat{\mathbf{y}}) = \frac{1}{n} \sum_{i=1}^{n} |y_{i} - \hat{y}_{i} |.
\end{equation*}
The NLPD is defined as the average negative log
value of the predictive distribution at test inputs $\mathbf{X}^{*}$ with test outputs
$\mathbf{y}^{*}$:
\begin{equation*}
  \text{NLPD}(\mathbf{y}^{*}) = -\frac{1}{n} \sum_{i=1}^{n} \log p_{\text{pred}}(y_{i}^{*} | \mathbf{x}_{i}^{*}).
\end{equation*}
This measure is commonly used to compare predictive performance on unseen data among
different models \citep{gelman2014understanding}.
For GPR-G and GPR-CN, $p_{\text{pred}}$ is given in
\cref{eq:aprox-pred-gaussian} and \cref{eq:pred-distr-scn}, respectively.
For GPR-t and GPR-L, $\log p_{\text{pred}}$ is approximated using Monte-Carlo methods, i.e.
\begin{equation*}
  \log p_{\text{pred}}(y^{*} | \mathbf{x}^{*}) \approx \frac{1}{M} \sum_{m=1}^{M} \log p(y^{*} | f^{(m)}, \vartheta), \ \  f^{(m)} \sim q(f^{(m)} | \mathbf{X}^{*}, \varphi),
\end{equation*}
where $M$ is the number of Monte-Carlo samples; $q$ is the variational distribution
given in \cref{eq:var-post-f}; and $\log p(y^{*} | f^{(m)}, \vartheta)$ is the
log-likelihood function for either the Student-t or Laplace distribution centered at
$f^{(m)}$ with hyperparameters $\vartheta$. We set $M = 1000$ when computing the NLPD
for GPR-t and GPR-L.
Furthermore, the expectation term in \cref{eq:elbo-factorized} for GPR-t and GPR-L are
also intractable and are approximated using the Gauss-Hermite quadrature method
\citep{liu1994note}. In each of the
simulations below, we use the popular squared exponential kernel with automatic
relevance determination (ARD) defined as
\begin{equation}
  k_{\text{se}}(\mathbf{x}, \mathbf{x}') = \sigma_{s}^{2} \exp \Bigg( - \sum_{j=1}^{p} \frac{|| x_{j} - x_{j}' ||^{2}}{2 \ell_{j}^{2}} \Bigg), \ \ \mathbf{x}, \mathbf{x}' \in \mathbb{R}^{p},
  \label{eq:squared-exp-kernel}
\end{equation}
where $\sigma_{s}^{2}$ is an output scale parameter; and $\{\ell_{j}\}_{j=1}^{p}$
are individual length scales for each input dimension.
We train each SVGPR model with 500 inducing points for 30 iterations using the Adam
algorithm implemented in PyTorch with an exponentially decaying step size starting at
0.1 and a batch size of 256 \citep{kingma2014adam,paszke2019pytorch}.
The inducing inputs were initialized to be a random sample of the training inputs.
Since the objective function for GPR-CN and GPR-t may have multiple local optima, we
rerun our algorithm five times with different initial parameter values and keep the run
that yields the largest value for \cref{eq:cn-elbo} in each simulation study.

\subsection{Parameter estimation}
\label{sec:parameter-estimation}

\begin{figure}[ht]
  \includegraphics[width=\columnwidth]{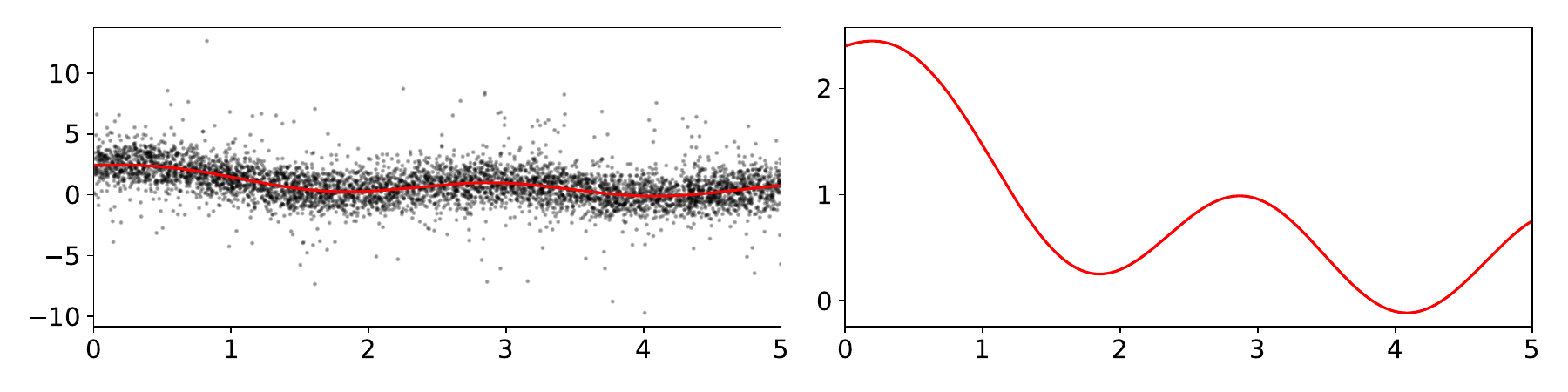}
  \caption{Simulated data for our first simulation study. The true function from
    eq.\ (\ref{eq:sim1-func}) is shown in red. Noisy observations are plotted on the
    \textbf{left}.}
  \label{fig:sim1-data}
\end{figure}

\begin{figure}[ht]
  \centering
  \includegraphics[width=\columnwidth]{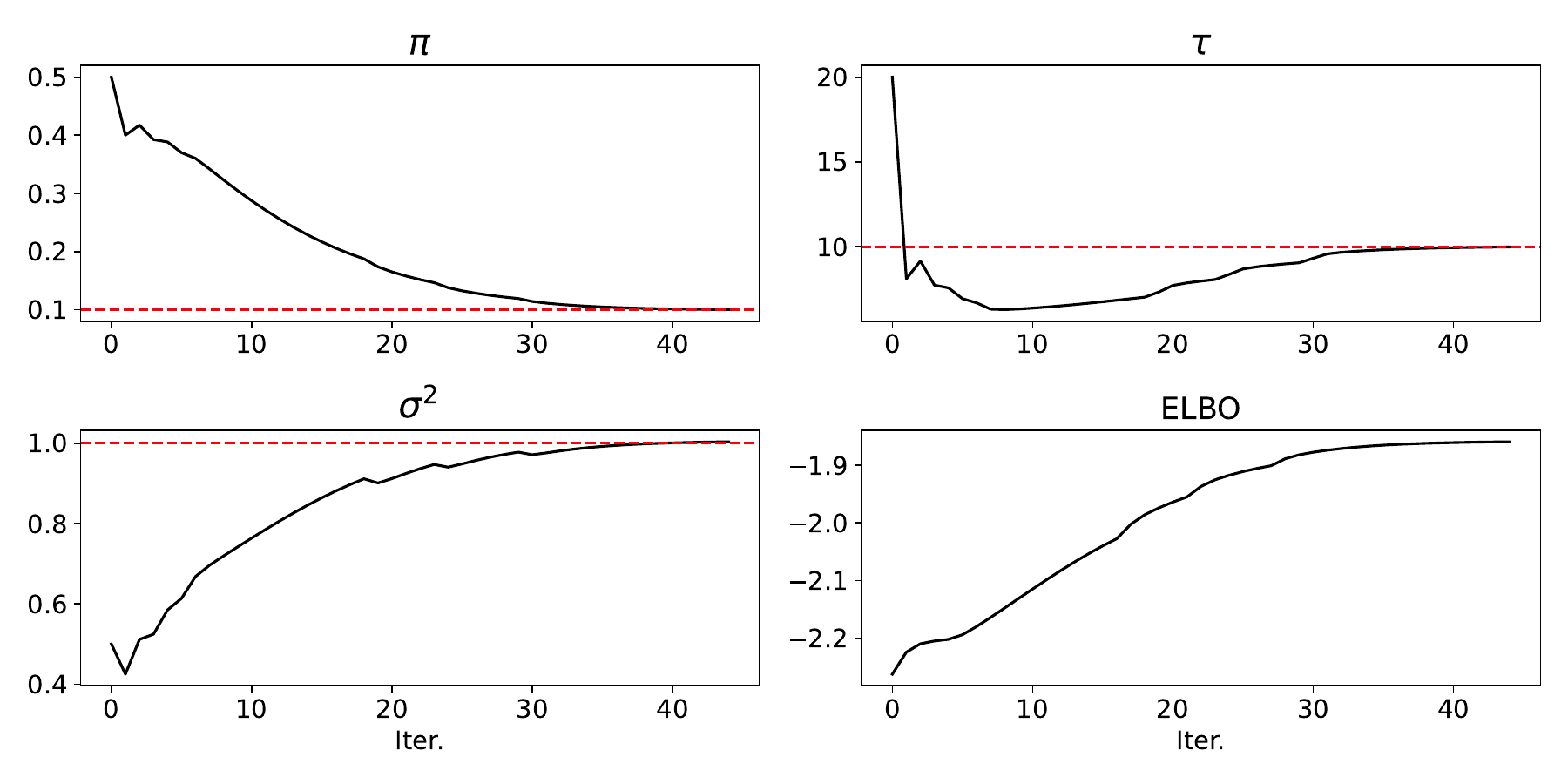}
  \caption{Parameter estimates and ELBO value across iterations for a specific run. Red
    dashed lines show the true parameter value.}
  \label{fig:sim1-param-traj}
\end{figure}

\begin{figure}[ht]
  \centering
  \includegraphics[width=\columnwidth]{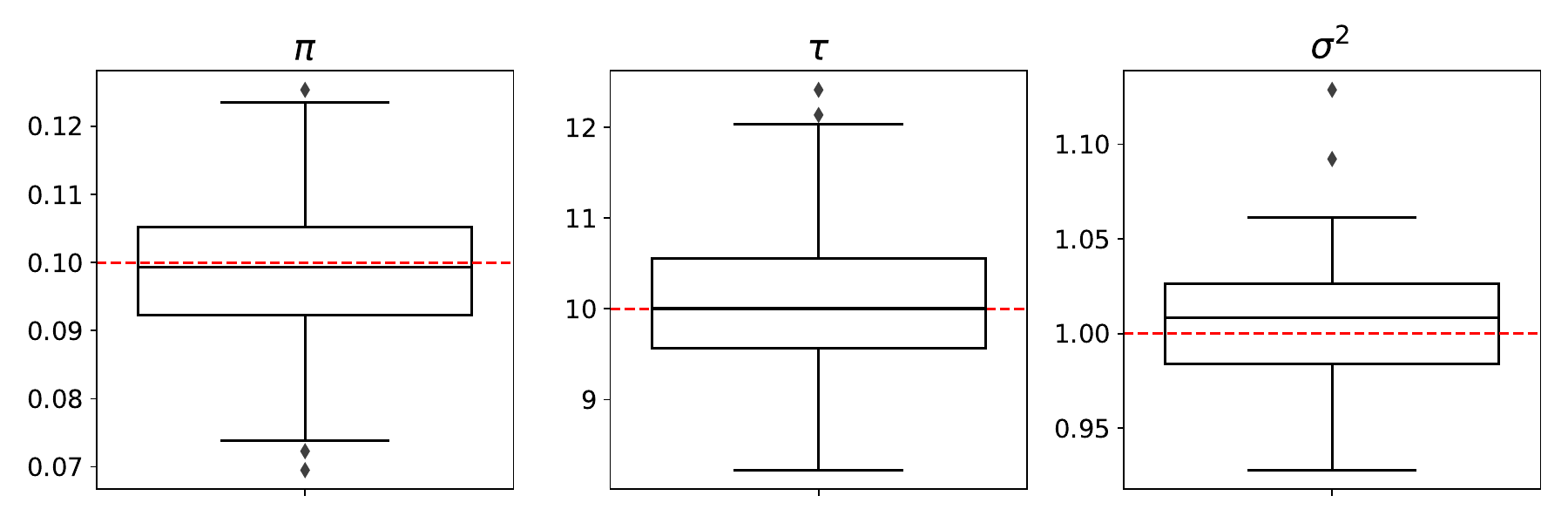}
  \caption{Boxplots for estimated parameter values from our proposed algorithm applied to 200
    different simulated datasets. Red dashed lines show the true parameter value.}
  \label{fig:sim1-boxplots}
\end{figure}

\begin{figure}[ht]
  \centering
  \includegraphics[width=\columnwidth]{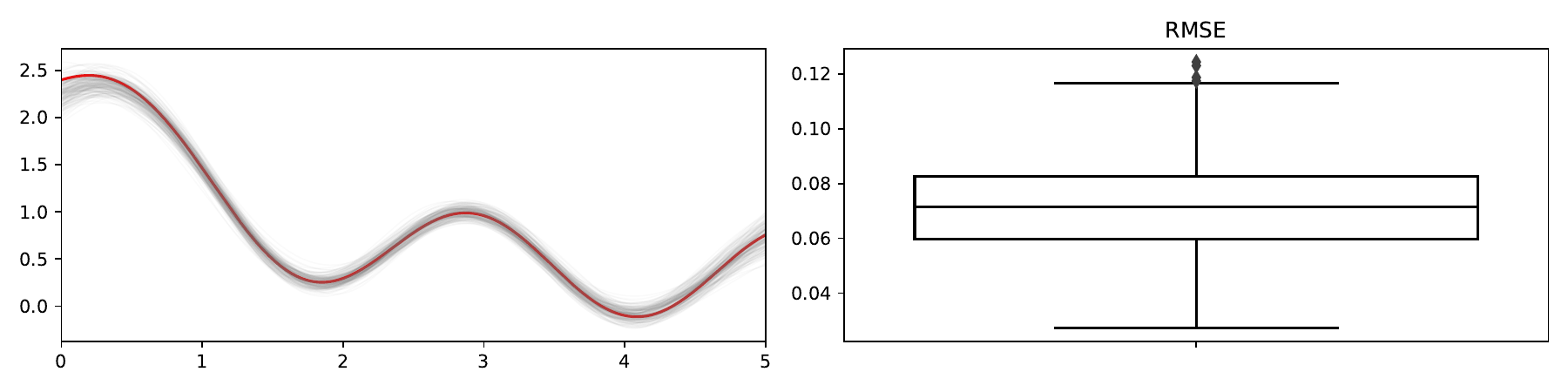}
  \caption{Function estimates from our proposed algorithm applied to 200 different
    simulated datasets. The true function from eq.\ (\ref{eq:sim1-func}) is shown in red.
    Estimated mean functions are shown in gray on the \textbf{left}. RMSEs for the
    estimated mean functions are given in the boxplot on the \textbf{right}.}
  \label{fig:sim1-func-est}
\end{figure}

For the first simulation study, we simulate 200 datasets of size $n = 5000$ directly from the
model in \cref{eq:scn-noise} with $\pi = 0.1$, $\tau = 10$, $\sigma^{2} = 1$, and
\begin{equation}Figures/
  f(x_{i}) = 0.3 + 0.4x_{i} + 0.5\text{sin}(2.7x_{i}) + \frac{1.1}{1 + x_{i}^{2}},
  \label{eq:sim1-func}
\end{equation}
where $x_{1}, \dots, x_{5000} \sim \text{Uniform}(0, 5)$. We plot one of the simulated
datasets in \cref{fig:sim1-data}. This was adapted from an artificial regression problem
considered by \cite{neal1997monte}. We plot the parameter estimates and corresponding
ELBO value across iterations for one dataset in \cref{fig:sim1-param-traj}. In this
specific run, the parameter estimates converge to a value close to the true value after
roughly 35 iterations. \Cref{fig:sim1-boxplots} shows boxplots for estimated parameter
values across the different simulated datasets. These boxplots show that most of the
estimated values are close to the true parameter values. We plot the estimated mean
functions and their respective RMSE values in \cref{fig:sim1-func-est}. From these
plots, we can see that the estimated mean functions are close to the true function most
of the time.

\subsection{Comparison with other robust likelihoods}
\label{sec:comp-to-sparse}

For the second simulation study, we consider an artificial
regression problem described in \cite{kuss2006gaussian} which uses the following
function first introduced in \cite{friedman1991multivariate}:
\begin{equation}
  f(\mathbf{x}) = 10\text{sin}(\pi x_{1} x_{2}) + 20(x_{3} - 0.5)^{2} + 10x_{4} + 5x_{5},
  \label{eq:sim2-func}
\end{equation}
where $\mathbf{x} = (x_{1}, \dots, x_{10})$. The last 5 dimensions in $\mathbf{x}$ are ignored in order to incorporate feature selection to the problem. We generate
$N_{\text{rep}} = 200$ datasets of size $n = 5000$ by sampling
$\mathbf{x}_{1}, \dots, \mathbf{x}_{5000}$ from the uniform distribution on the unit
hyper-cube $[0, 1]^{10}$. We then compute the corresponding function values in
\cref{eq:sim2-func} and add standard normal noise to it, i.e.
\begin{equation*}
  y_{i} = f(\mathbf{x}_{i}) + \epsilon_{i}, \ \ \epsilon_{i} \sim N(0, 1).
\end{equation*}

Lastly, we add outliers by replacing $p_{\text{outlier}}$ proportion of the generated observations with samples
drawn from $N(15, \sigma_{\text{outlier}}^{2})$, where $p_{\text{outlier}} \in \{0.1, 0.2, 0.3\}$ and $\sigma_{\text{outlier}} \in \{3, 10\}$.
In the $\sigma_{\text{outlier}} = 3$ case, the generated outliers are likely to lie in
the same range as the function values. Setting $\sigma_{\text{outlier}} = 10$
constitutes a more difficult case where the outliers are unrelated to the function and are
likely to lie outside of the function value range. We consider four scenarios with
increasing outlier proportion and magnitude to study the performance of the considered
noise models. These scenarios are summarized in \cref{tbl:sim2-scenarios}. The first
scenario is the same one considered in \cite{kuss2006gaussian} and serves as a baseline
for the remaining scenarios which have an increasing proportion of extreme outliers.
We fit the various GPR models to simulated datasets generated under various outlier
scenarios and evaluate them on 10,000 noise-free test samples of \cref{eq:sim2-func}.
\begin{table}[ht]
  \centering
  \begin{tabular}{l c c c}
    \toprule
    & $p_{\text{outlier}}$ & $\sigma_{\text{outlier}}$ & Description \\
    \midrule
    1 & 0.1 & 3 & Low proportion, mild outliers \\
    2 & 0.1 & 10 & Low proportion, extreme outliers \\
    3 & 0.2 & 10 & Medium proportion, extreme outliers \\
    4 & 0.3 & 10 & Large proportion, extreme outliers \\
    \bottomrule
  \end{tabular}
  \caption{Summary of outlier scenarios considered in the second simulation study.}
  \label{tbl:sim2-scenarios}
\end{table}

\begin{figure*}[ht!]
  \centering
  \begin{subfigure}[b]{\columnwidth}
    \includegraphics[width=\columnwidth]{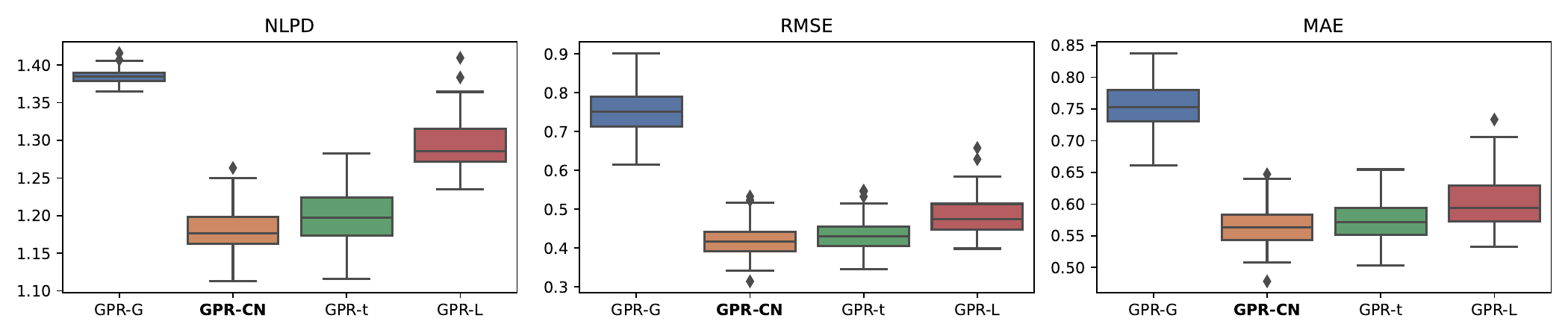}
    \caption{$p_{\text{outlier}} = 0.1$, $\sigma_{\text{outlier}} = 3$}
    \label{fig:sim2-easy}
  \end{subfigure}
  \begin{subfigure}[b]{\columnwidth}
    \includegraphics[width=\columnwidth]{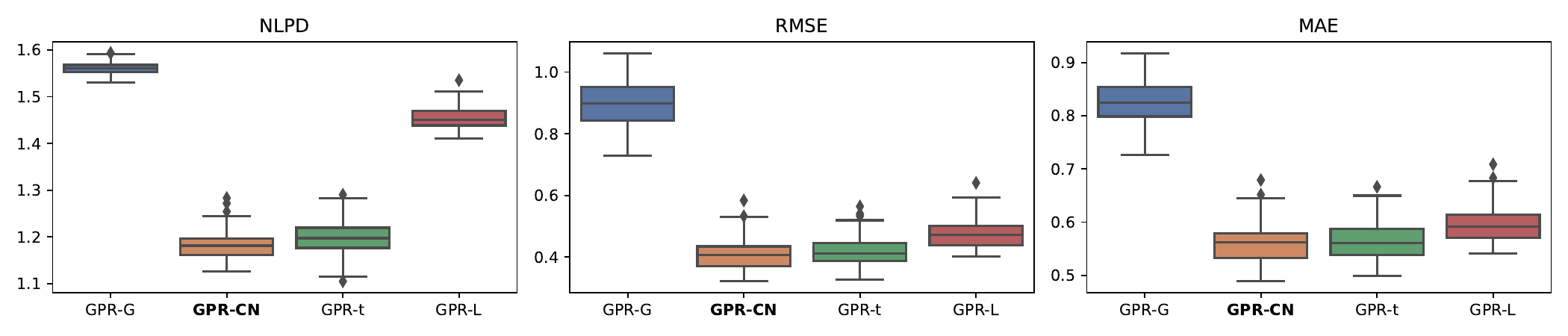}
    \caption{$p_{\text{outlier}} = 0.1$, $\sigma_{\text{outlier}} = 10$}
    \label{fig:sim2-med}
  \end{subfigure}
  \begin{subfigure}[b]{\columnwidth}
    \includegraphics[width=\columnwidth]{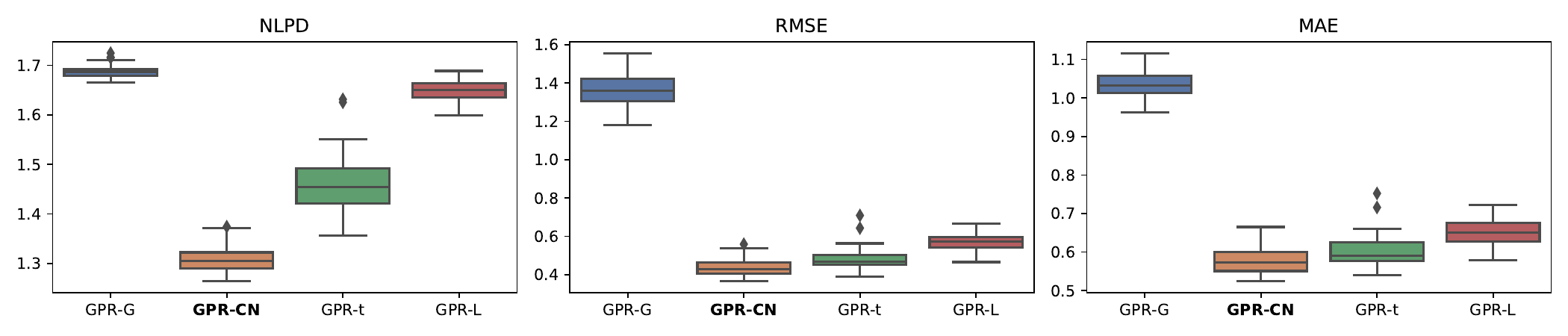}
    \caption{$p_{\text{outlier}} = 0.2$, $\sigma_{\text{outlier}} = 10$}
    \label{fig:sim2-med-hard}
  \end{subfigure}
  \begin{subfigure}[b]{\columnwidth}
    \includegraphics[width=\columnwidth]{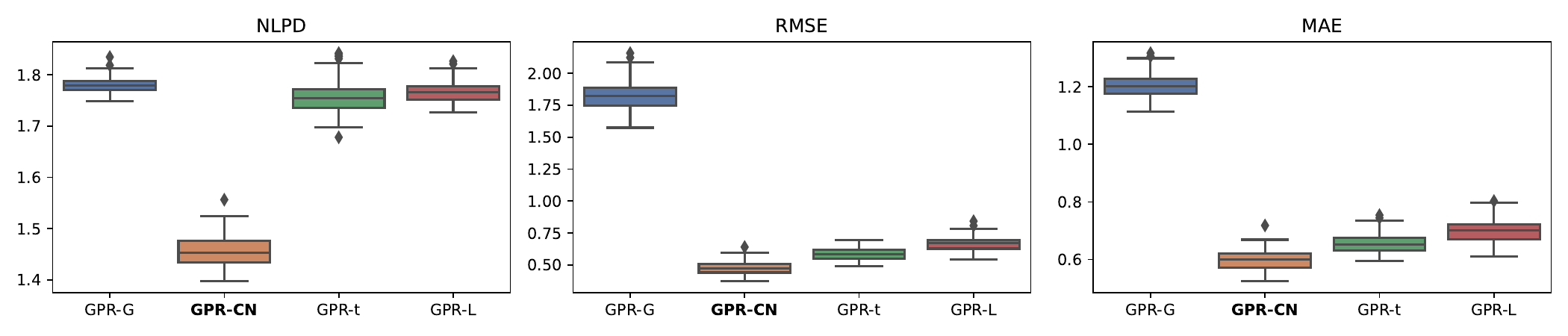}
    \caption{$p_{\text{outlier}} = 0.3$, $\sigma_{\text{outlier}} = 10$}
    \label{fig:sim2-hard}
  \end{subfigure}
  \caption{Boxplots of NLPD, RMSE, and MAE for second simulation study with varying
    $p_{\text{outlier}}$ and $\sigma_{\text{outlier}}$ across 200 simulated datasets.}
  \label{fig:sim2-boxplots}
\end{figure*}

Boxplots of the RMSE, MAE, and NLPD for the different models and outlier scenarios are
given in \cref{fig:sim2-boxplots}. Note that the boxplots for GPR-CN and GPR-t tend to
have a higher spread compared to GPR-G and GPR-L. This is likely because these methods
have ELBOs that are not log-concave and are vulnerable to local optima. Furthermore, the
NLPDs for GPR-t are approximated with Monte-Carlo samples.
In all scenarios, the robust noise models (GPR-CN, GPR-t, GPR-L) outperform GPR-G in
terms of all metrics. The RMSEs and MAEs are similar for the robust noise models
(GPR-CN, GPR-t, GPR-L) in all scenarios.
In the first scenario, the NLPDs are similar for GPR-CN and GPR-t but are
slightly higher for GPR-L. This difference in NLPDs is exacerbated in the remaining
scenarios, where the outliers are more extreme. This is likely because the Laplace
distribution is not as heavy-tailed as the t-distribution and does not model outliers
separately as in the CN distribution.
As we increase the proportion of outliers, GPR-CN starts to outperform GPR-t in terms of NLPD.
This suggests that GPR-CN is more adept at handling larger proportions of outliers than
GPR-t. This may be due to the fact that outliers are identified and placed into its own
separate component in GPR-CN so they don't affect model fit for non-outliers. On the
other hand, outliers are not explicitly isolated in GPR-t so they may affect the fit for
non-outliers. As the proportion of outliers increases, we may expect the true latent
function posterior to be multimodal, making the Gaussian variational distribution
inadequate for approximating it. However, this does not seem to hinder predictive
performance as can be seen from the results in the third and fourth scenario.

\section{Applications}
\label{sec:applications}

In this section, we compare the numerical performances of the proposed algorithm with state-of-the-art literature on two data sets. One is the flight delay data, which has been adopted extensively in the Gaussian process literature, and the other is the ground magnetic perturbation data that we described in the Introduction.  The flight delay data serves as a real-data evaluation of our model, and the ground magnetic perturbation data is of great scientific interest. We fit the four SVGPR models identified in
\cref{sec:comp-to-sparse} on flight delays. For ground magnetic perturbations data ($\delta B_H$), we fit a
dense artificial neural network and a GPR-CN model and compare their predictive performances with skill scores, RMSE, and focus in depth
on the invterval coverage properties.

The model fitting procedure varies slightly between datasets. For the flights data,
we train the four GPR models with 1000 inducing points for up to 30 iterations using the Adam algorithm implemented in
PyTorch with an exponentially step size starting at 0.1 and a batch size of 256
\citep{kingma2014adam,paszke2019pytorch}. The inducing inputs were initialized to be a random sample of the training inputs. We used a validation set to monitor the
validation NLPD and terminate training if it hasn't decreased in 5 iterations. Input
features are standardized by subtracting the training sample mean and dividing by the
training sample standard deviation.

For $\delta B_H$ prediction, we use a more flexible hyperparameter setup, opting to allow the number of inducing points and batch size to vary and choosing optimal hyperparameters using the Bayesian Optimization package \cite{bayesopt}. 
As the data is fully numeric and relatively small station by station,
we use a large batch size on a GPU to speed up computation. The details of this can be found in \ref{sec:model-fits}.

For the flight delays data, we use the ARD squared exponential
kernel defined in \cref{eq:squared-exp-kernel}. For the ground magnetic perturbations data, we use the ARD Matern kernel with smoothness parameter $\nu$ defined as
\begin{equation*}
  k_{\text{matern}}(\mathbf{x}, \mathbf{x}') = \sigma_{s}^{2} \frac{2^{1-\nu}}{\Gamma(\nu)} \big( \sqrt{2\nu} d \big)^{\nu} K_{\nu} \big(\sqrt{2 \nu} d \big), \ \ \mathbf{x}, \mathbf{x}' \in \mathbb{R}^{p},
\end{equation*}
where $d = \sum_{j=1}^{p} \ell_{j}^{-2} ||x_{j} - x_{j}'||^{2}$; $K_{\nu}$ is the modified
Bessel function; $\sigma_{s}^{2}$ is an output scale parameter; and
$\{\ell_{j}\}_{j=1}^{p}$ are individual length scales for each input dimension. We set
$\nu = 1.5$ which allows us to rewrite $k_{\text{matern}}$ as
\begin{equation*}
  k_{\text{matern}}(\mathbf{x}, \mathbf{x}') = \sigma_{s}^{2} \big( 1 + \sqrt{3}d \big) \exp \big( - \sqrt{3} d\big).
\end{equation*}

\subsection{Flight delays}
\label{sec:flight-delays}

\begin{figure}[ht]
  \centering
  \includegraphics[width=\columnwidth]{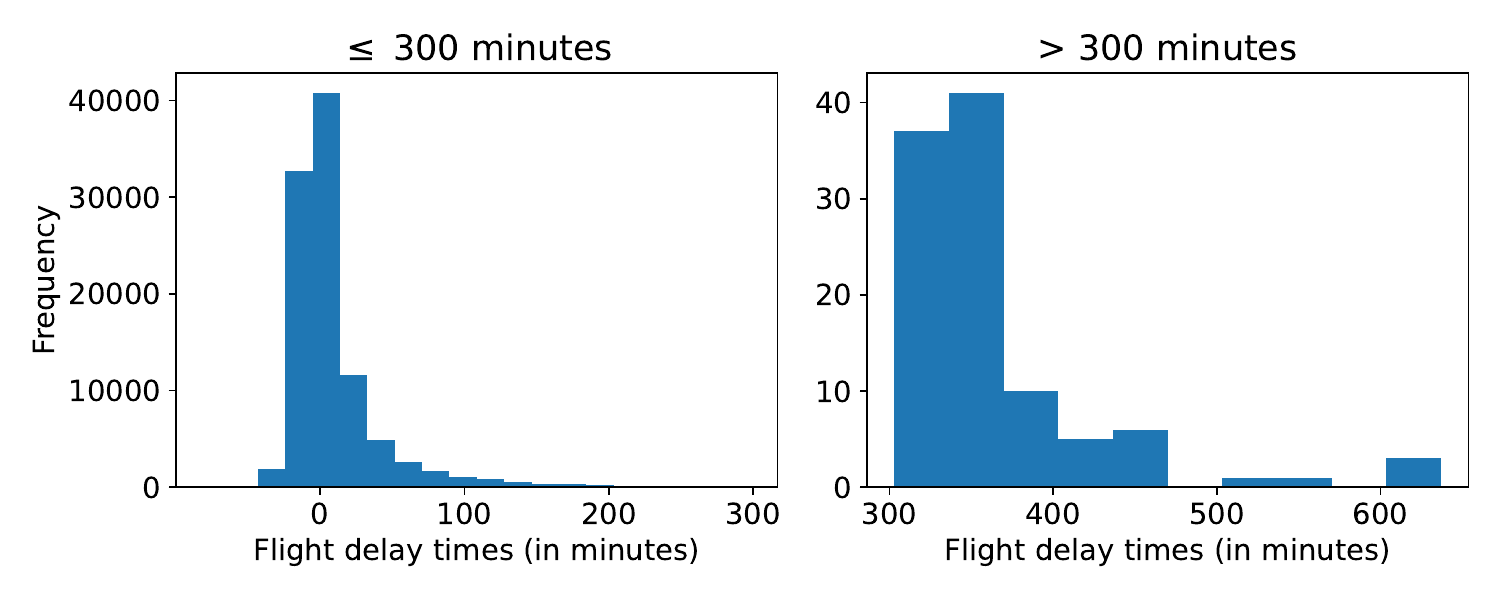}
  \caption{Histograms of flight delay times (in minutes) used for training. Left plot shows
    histogram for delays less than or equal to 300 minutes. Right plot shows histogram for delays
    greater than 300 minutes.}
  \label{fig:delays-hist}
\end{figure}

The flight delays dataset
consists of information about commercial flights in the US from January 2008 to April
2008 and was taken from an example in \cite{hensman2013gaussian}. It has become a standard
benchmark dataset in the GPR literature for comparing scalable GPR methods due to its
massive size and non-stationary nature. The original dataset curated
by \cite{hensman2013gaussian} contains 800,000 observations that were randomly sampled
from a dataset containing around 2 million observations. To speed up computation, we
reduced the size by randomly sampling 100,000 observations for training, and 20,000 each
for validating and testing.
The goal is to predict flight delay times (in minutes) using the eight features considered by
\cite{hensman2013gaussian}: aircraft age, distance to travel, airtime, departure time,
arrival time, day of the week, day of the month, and month. A histogram of flight delay
times used for training is given in \cref{fig:delays-hist}.
99\% of flights are only delayed for up to 166 minutes. However, there are a handful of
flights that are delayed for more than 300 minutes. These delays are likely due to
external factors such as weather that are not considered in our models. This motivates the use of robust models for analyzing this data.



\begin{table}[ht]
  \centering
  \begin{tabular}{lllllll}
    Min. & Q25 & Median & Mean & Q75 & Q99 & Max \\
  \toprule
    -81 & -9 & -1 & 9.4 & 13 & 166 & 637 \\
  \bottomrule
  \end{tabular}
  \caption{Summary statistics for flight delay times (in minutes) used for training.}
\end{table}

We fit the
models considered in \cref{sec:comp-to-sparse} to this dataset and compare their
predictive performances.
For this dataset, we also use the ARD squared exponential kernel defined in \cref{eq:squared-exp-kernel}.
The NLPD, RMSE, and MAE for the four GPR models fitted to this dataset are
reported in \cref{tbl:flight-delays-res}. GPR-CN has the lowest NLPD and ties with GPR-L
for the lowest MAE. GPR-G has the lowest RMSE but highest MAE. This suggests that the
resulting model fit for GPR-G is more heavily influenced by outliers than the ones for
the other methods.

\begin{table}[ht]
  \centering
  \begin{tabular}{l|c|c|c}
    \toprule
    & \textbf{NLPD} & \textbf{RMSE} & \textbf{MAE} \\
    \midrule
    \textbf{GPR-G} & 4.95 & \textbf{34.38} & 4.65 \\
    \textbf{GPR-CN} & \textbf{4.51} & 36.94 & \textbf{4.32} \\
    \textbf{GPR-t} & 4.62 & 37.55 & 4.35 \\
    \textbf{GPR-L} & 5.38 & 36.12 & \textbf{4.32} \\
    \bottomrule
  \end{tabular}
  \caption{Metrics for GPR models trained on flight delays data. The lowest value(s) in each column are shown in \textbf{bold}.}
  \label{tbl:flight-delays-res}
\end{table}

\subsection{Ground magnetic perturbations}
\label{sec:space-weath-forec}

To determine the effectiveness of the GPR-CN on real world data, we turn our attention back to $\delta B_H$ prediction. This data differs 
from the flights dataset considerably; first, it is messier, and more prone to missingness and measurement error. 
We examine the results of applying the GPR-CN model to $\delta B_H$ prediction below.

\subsubsection{Data Setup}

For this project, we chose twelve test stations commonly used in the space weather
literature\footnote{These stations are: Yellowknife (YKC), Meanook (MEA), Newport (NEW), Fresno (FRN), Iqaluit (IQA), Poste de la Baleine (PBQ), Ottawa (OTT), Fredericksburg (FRD), Hornsund (HRN), Abisko (ABK), Wingst (WNG), and Furstenfeldbruk (FUR).}. 

For the FUR station, 
we were unable to produce a consistent GPR-CN fit, and so the results are 
reported for the remaining eleven. Due to missing data, the station PBQ is 
replaced by the nearby station T31/SNK (Sanikiluaq); the 
FRN station (Fresno) is replaced by the nearby station T16 (Carson City) 
for the year 2015.

\begin{table}
\caption{Summary statistics of GPR-CN and ANN models for the testing period of 2015-2016. Target coverage is 95\%. Coverage is comparable in both storm and non-storm periods, but the IQR of interval length is generally either comparable or lower in the GPR-CN model.}
\label{tbl:station-perf}
 
\begin{tabular*}{\linewidth}{@{\extracolsep{\fill}} lllllll }
\toprule
 \multicolumn{7}{c}{All Test Data} \\ 
& \multicolumn{2}{l}{RMSE} & \multicolumn{2}{l}{Coverage} & \multicolumn{2}{l}{Interval Length (IQR)} \\
Model & ANN & GPR-CN & ANN & GPR-CN & ANN & GPR-CN  \\
Station & &  &  &  &  &\\
\midrule
ABK & 87.874 & 86.055 & 0.934 & 0.908 & 322.678 & 104.563 \\
FRD & 11.861 & 13.044 & 0.938 & 0.978 & 20.813 & 29.105 \\
FRN & 14.012 & 15.558 & 0.934 & 0.963 & 21.230 & 30.694 \\
HRN & 65.169 & 65.712 & 0.941 & 0.946 & 164.701 & 121.076 \\
IQA & 81.283 & 83.912 & 0.894 & 0.979 & 207.880 & 237.874 \\
MEA & 73.432 & 78.744 & 0.929 & 0.909 & 68.185 & 40.418 \\
NEW & 16.420 & 18.681 & 0.923 & 0.952 & 24.106 & 25.724 \\
OTT & 15.522 & 18.473 & 0.918 & 0.941 & 21.987 & 25.464 \\
PBQ & 85.699 & 92.237 & 0.934 & 0.913 & 171.892 & 108.511 \\
WNG & 13.923 & 15.007 & 0.925 & 0.941 & 23.780 & 23.911 \\
YKC & 93.764 & 96.636 & 0.948 & 0.951 & 287.425 & 208.491 \\
\bottomrule

\end{tabular*} 
\begin{tabular*}{\linewidth}{@{\extracolsep{\fill}} lllllll }
\toprule
\multicolumn{7}{c}{Storm Periods Only} \\ 
& \multicolumn{2}{l}{RMSE} & \multicolumn{2}{l}{Coverage} & \multicolumn{2}{l}{Interval Length (IQR)} \\
Model & ANN & GPR-CN & ANN & GPR-CN & ANN & GPR-CN \\
Station &  &  &  &  &  &\\
\midrule
ABK & 145.126 & 118.050 & 0.939 & 0.905 & 891.077 & 167.798 \\
FRD & 25.133 & 31.673 & 0.883 & 0.964 & 41.961 & 47.423 \\
FRN & 33.742 & 36.837 & 0.840 & 0.890 & 44.144 & 46.963 \\
HRN & 104.870 & 106.009 & 0.878 & 0.901 & 333.435 & 204.989 \\
IQA & 109.992 & 128.926 & 0.966 & 0.951 & 320.797 & 295.607 \\
MEA & 114.532 & 132.416 & 0.933 & 0.870 & 148.557 & 68.892 \\
NEW & 33.443 & 41.519 & 0.921 & 0.931 & 43.032 & 45.698 \\
OTT & 40.779 & 55.719 & 0.846 & 0.863 & 36.850 & 40.187 \\
PBQ & 113.425 & 129.959 & 0.933 & 0.901 & 296.378 & 189.384 \\
WNG & 29.586 & 35.025 & 0.885 & 0.893 & 38.889 & 37.622 \\
YKC & 144.834 & 158.726 & 0.950 & 0.942 & 482.269 & 319.620 \\
\bottomrule
\end{tabular*} 

\end{table}

For each of these twenty minute intervals, we use lagged OMNI data as our 
primary input features. The OMNI data is collected every minute, but we use the median 
value for every five minutes over the previous hour as our input features. After this, 
we compute seasonal features for both time of day and the day of the year. 
This gives us  a total of ninety-six input features with which to predict 
$\delta B_H$ for each station.

Following the convention made in the space weather community, we do not include lagged DeltaB values as inputs \citep{keesee2020comparison,upendran2022global,pinto2022revisiting}.
Therefore, we will ignore the temporal nature of this data and treat it as a
conventional regression problem.
Space physicists are typically most
interested in the predictive performance during geomagnetic storms when $\delta B_{H}$ spikes, and so
we report results across the entire prediction range as well as within storm performance.

The SuperMAG data is stored in Parquet files divided by station. The data are loaded
from these files and any missing points are dropped; as our approach is to use
covariates only and not any lagged SuperMAG data, gaps in the response variable
do not significantly affect the model fit. Once we have a SuperMAG series
with no missing entries, relevant OMNI features are matched by timestamp to 
the SuperMAG series.

\subsubsection{Models}
\label{sec:model-fits}

To fit the models, we divided the data into three time periods. Training consists of
the twenty minute maximum $\delta B_H$ value from 2010-2013. We held out 2014 and 
2015 as validation and testing sets, respectively. Model fits are 
performed independently by station. In addition to the GPR-CN model, we fit a 
dense ANN to serve as a performance baseline.

In the context of ANN model, deriving a probabilistic interpretation of uncertainties and contrasting this with GPR methods poses a unique challenge. To facilitate inference within our predictions, we assume that each data point follows a normal distribution, and we employ a 4-layer deep multi-output neural network designed to estimate both location parameters $\hat{\mu}$ and scale parameters $\hat{\sigma}^2$. This approach is underpinned by a customized loss function $L_{\mathrm{ANN}}$, that combines RMSE and NLPD with a fixed weight $\lambda$. Consequently, the loss function for our ANN model is defined as
$$
L_{\mathrm{ANN}}(\mathbf{y}, \hat{\mathbf{\mu}}, \hat{\mathbf{\sigma}}^2)=\operatorname{RMSE}(\mathbf{y}, \hat{\mathbf{\mu}})+\lambda \operatorname{NLPD}(\mathbf{y} \mid \hat{\mathbf{\mu}}, \hat{\mathbf{\sigma}}^2) .
$$
This framework enables the construction of a $95 \%$ prediction interval $[\hat{Z}_{0.025}, \hat{Z}_{0.975}]$, where $\hat{Z}$ is distributed normally with parameters $\hat{\mu}$ and $\hat{\sigma}^2$. The disadvantage of the ANN model comparing to the GPR-CN model is that the combined loss function is artificial, as it balances two different aspects of model performance. We have to tune an additional hyperparameter $\lambda$ and ensure the convergence of NLPD loss. The detailed procedure of hyperparameter tuning is included in \cref{sec:annfit}.

The GPR-CN model is fit using the ARD Matern kernel with smoothness parameter $\nu$ defined as
\begin{equation*}
  k_{\text{matern}}(\mathbf{x}, \mathbf{x}') = \sigma_{s}^{2} \frac{2^{1-\nu}}{\Gamma(\nu)} \big( \sqrt{2\nu} d \big)^{\nu} K_{\nu} \big(\sqrt{2 \nu} d \big), \ \ \mathbf{x}, \mathbf{x}' \in \mathbb{R}^{p},
\end{equation*}
where $d = \sum_{j=1}^{p} \ell_{j}^{-2} ||x_{j} - x_{j}'||^{2}$; $K_{\nu}$ is the modified
Bessel function; $\sigma_{s}^{2}$ is an output scale parameter; and
$\{\ell_{j}\}_{j=1}^{p}$ are individual length scales for each input dimension. We set
$\nu = 1.5$ which allows us to rewrite $k_{\text{matern}}$ as
\begin{equation*}
  k_{\text{matern}}(\mathbf{x}, \mathbf{x}') = \sigma_{s}^{2} \big( 1 + \sqrt{3}d \big) \exp \big( - \sqrt{3} d\big).
\end{equation*}

We use only the single Matern kernel, as opposed to additive or multiplicative kernel
constructions. This is a simple kernel setup, but works sufficiently well in practice. 

To fit the GPR-CN models, we use a Linux server with 32GB of RAM, a 16-core 
processor, and a 4GB NVidia GeForce 2060 GPU. As with the ANN, we use the 
Bayesian-Optimization library \cite{bayesopt} to find optimal hyperparameters. While the GPU is not 
technically necessary, it should be noted that hyperparameter tuning
on the CPU is a lengthy process, and GP models can be susceptible to the 
starting points of tuning parameters. The GPU enables running many more 
iterations of both model fitting and hyperparameter tuning, though it comes at a cost of model 
flexibility, namely by limiting the kernel setup. Due to memory constraints 
on the GPU, a periodic component could not be added to the existing 
kernel, and instead, seasonal
features were added to the data to remedy this. Next, GPU training adds
some complexity to the code primarily around data-type handling, though this 
is mitigated by the robust support of both the PyTorch and GPyTorch packages. 

Though the GPU-training is optional, it vastly improved the number of 
iterations we were able to perform for both hyperparameter training and model testing, 
and is therefore recommended for any extensions to this research.

\subsubsection{Results}

In terms of accuracy, the ANN model performs slightly better than the GPR-CN model.
Overall RMSE is lower on most stations, though both models are comparable. In addition to RMSE, we 
calculate Heidke and True Skill scores using station-specific deciles, as in
\cite{camporeale2020gray}.

\begin{figure}
\centering
{\includegraphics[width=0.7\columnwidth]{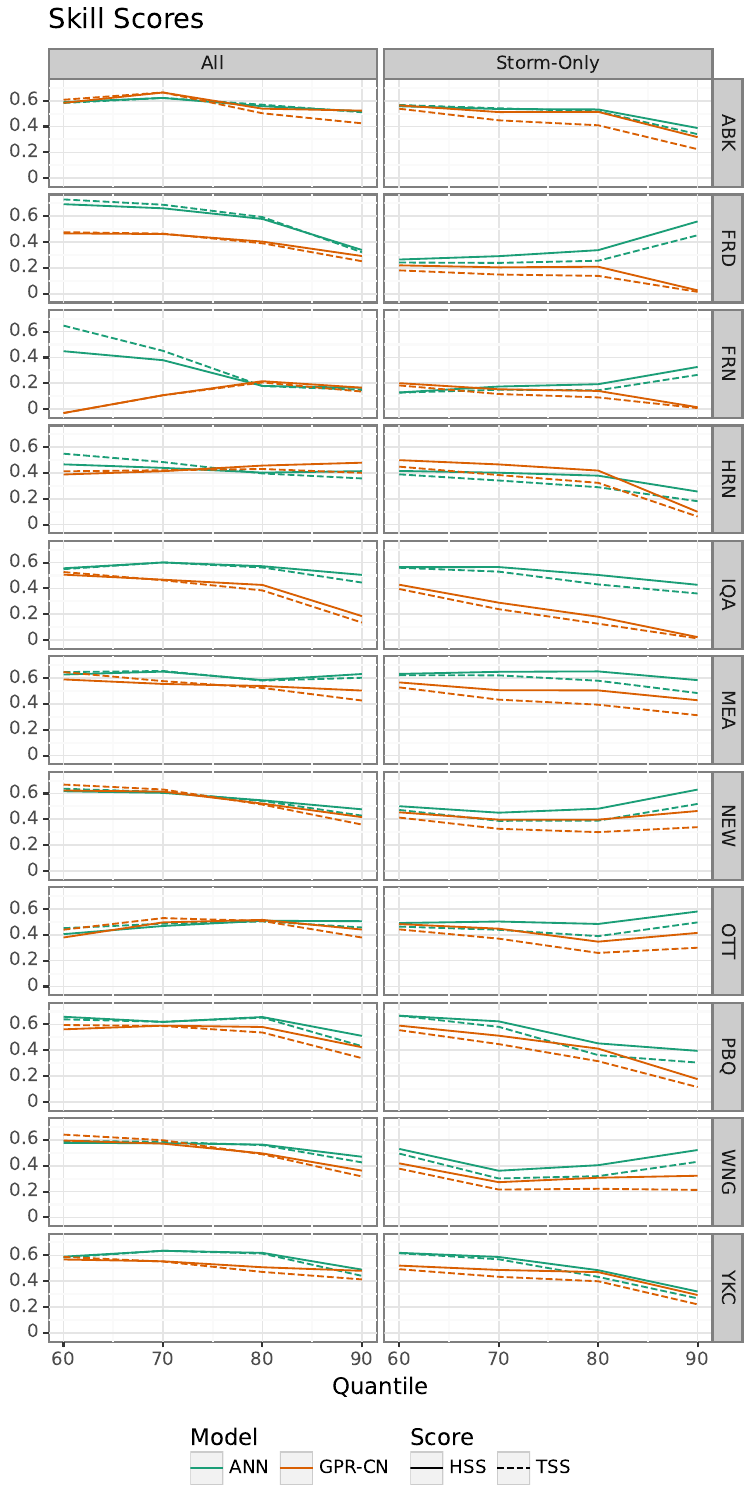}}
\label{fig:skill-scores}
\caption{Skill scores for each method by station. The quantiles on the x-axis correspond to the station's $\delta B_H$ quantile over the training period. These scores represent how well a model does at predicting when $\delta B_H$ will exceed the stated quantile. Higher is better, and the theoretical max is 1.}
\end{figure}

Differences in both the Heidke skill scores (HSS) and true skill scores (TSS) 
scores follow the same pattern as RMSE (see \cref{fig:skill-scores}). In general, the GPR-CN model is 
more conservative than the ANN, and as a result has a harder time predicting the extremes of the 
$\delta B_H$ distributions. Though skill scores for the GPR-CN and the ANN show the same trend 
in most stations, the gap in score between the ANN and GPR-CN methods increases with the 
percentile of $\delta B_H$ under consideration.

\subsubsection{Discussion}

The GPR-CN model is a more conservative approach to modelling 
than the ANN. On average, the GPR-CN model under predicts when
compared to the ANN. As a result, the conservatism of the GPR-CN model causes it to have diminished accuracy compared to the ANN across both the variance and bias of the errors (see \cref{tbl:bias_var_tradeoff} in the appendix).

While this conservatism leads to lower accuracy, it yields better estimates of the variance for a given prediction. The ANN often overestimates the standard deviation of a given prediction to the point of 
creating arbitrarily large confidence intervals that achieve coverage by guessing an impossibly large range. This problem is 
especially pronounced in storm periods, but the GPR-CN method largely avoids this issue. Though the GPR-CN model yields worse accuracy, the largest errors are considerably smaller than the largest errors from the ANN.
Across the majority of stations tested, the GPR-CN model has a much smaller 95th and 99th percentile of error, especially 
looking at high latitude stations (see \cref{tbl:error_percentiles}).

We have considered other uncertainty quantification methods such as MC dropout from \cite{gal2016dropout}. However, our data is a time series of extreme events, which introduces significant aleatoric (data) uncertainty due to high variability. MC Dropout primarily models epistemic (model) uncertainty and does not effectively capture aleatoric uncertainty. Even with an increased dropout rate of 0.7, the coverage rate with MC dropout remains poor (see \cref{tab:ANN_MC-Dropout}). By incorporating negative log-likelihood loss, our method models both the mean and variance, directly measuring aleatoric uncertainty. 

\begin{table}
\centering
\caption{95th and 99th percentile of errors for the ANN and GPR-CN models. GPR-CN consistently has lower errors at the top of the range for all stations but one.}
\label{tbl:error_percentiles}
\begin{tabular}{lllll}
\toprule
 & \multicolumn{2}{l}{Q95} & \multicolumn{2}{l}{Q99} \\
Model & ANN & GPR-CN & ANN & GPR-CN \\
Station &  &  &  &  \\
\midrule
ABK & 6378.09 & 469.85 & 9948.65 & 1186.73 \\
FRD & 115.44 & 110.04 & 518.18 & 213.88 \\
FRN & 100.98 & 119.79 & 344.42 & 234.25 \\
HRN & 918.05 & 433.00 & 4288.03 & 713.55 \\
IQA & 1048.26 & 795.39 & 4367.02 & 1137.49 \\
MEA & 796.80 & 200.42 & 7301.83 & 564.96 \\
NEW & 121.76 & 99.35 & 480.85 & 193.53 \\
OTT & 112.96 & 96.77 & 446.79 & 175.52 \\
PBQ & 1794.57 & 525.56 & 8331.53 & 1141.94 \\
WNG & 110.40 & 90.96 & 390.30 & 165.83 \\
YKC & 1724.41 & 868.48 & 6991.68 & 1577.43 \\
\bottomrule
\end{tabular}
\end{table}

\begin{figure}[!htb]
\centering
{\includegraphics[width=\columnwidth]{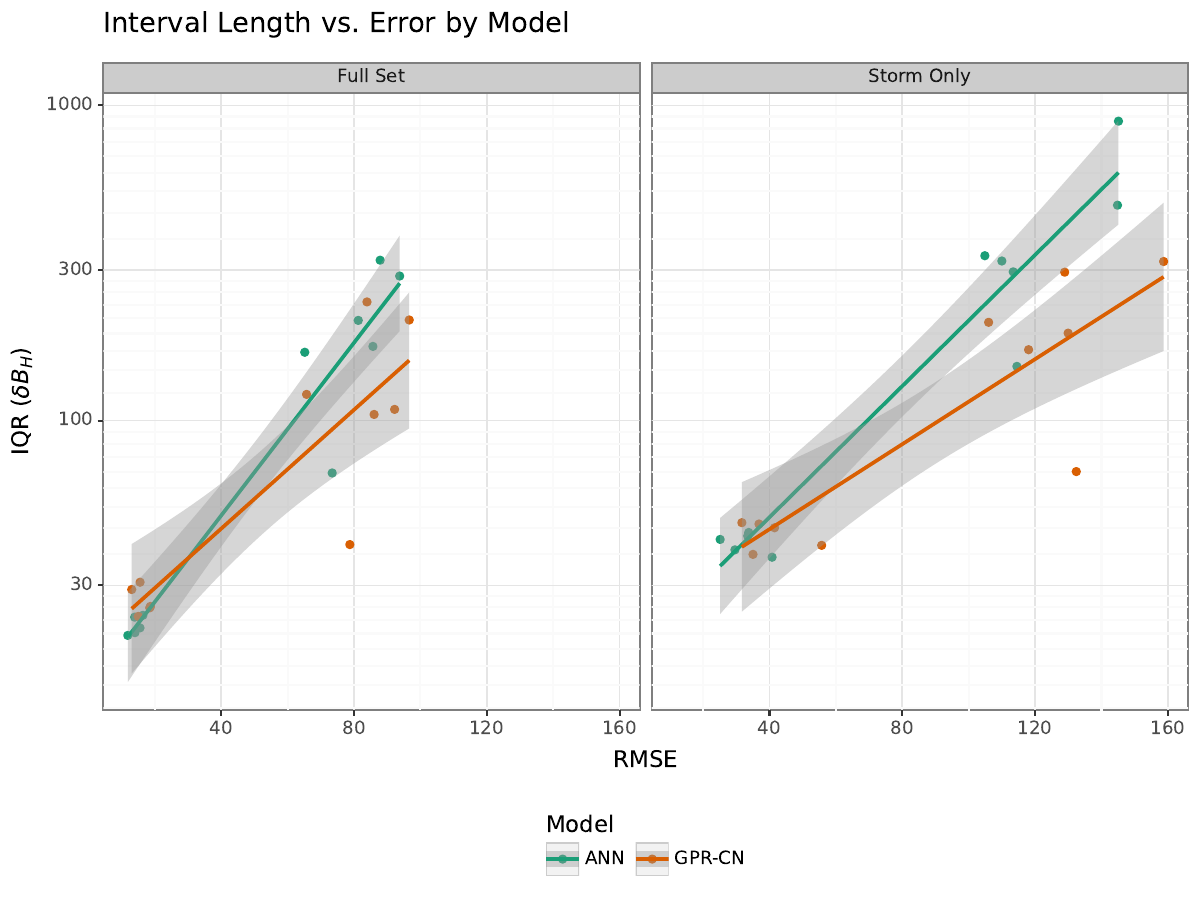}}
\caption{OLS regression on the IQR of prediction interval lengths vs. RMSE by model. Among the high RMSE stations, the IQR of prediction intervals during storms is lower for the GPR-CN model compared to the ANN. Note that the y-axis is visually log-scaled, though the units are preserved on the original scale.}
\label{fig:rmse-ols}

\end{figure}

\begin{figure}[!htb]

\centering
{\includegraphics[width=\columnwidth]{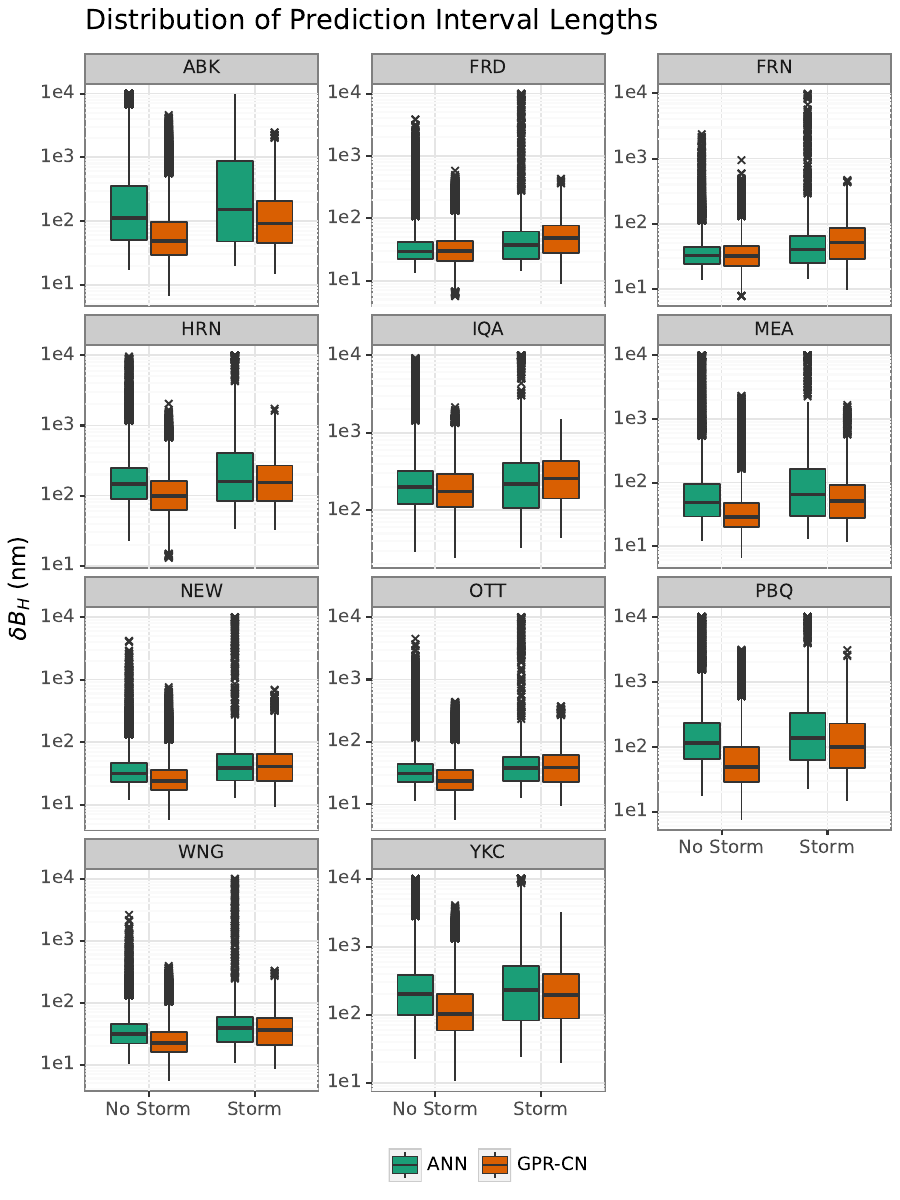}}
\caption{Distribution of interval length for storm and non-storm periods by method.}
\label{fig:interval-length}

\end{figure}

The primary advantage of the GPR-CN model is the interval length. Though the GPR-CN model 
sacrifices accuracy when compared to the ANN, it does so for the benefit of having better constrained
prediction intervals. For nine stations of eleven, the median interval length was lower than the ANN, and in all stations, 
the GPR-CN model had lower maximum interval lengths for similar coverage (\cref{tbl:station-perf}, \cref{fig:interval-length}). Further, 
as the RMSE of a particular station increases, the IQR of interval lengths increases as well, but this increase is less severe for the GPR-CN model (see \cref{fig:rmse-ols}).
This improvement to confidence intervals is especially pronouned examining the high RMSE (re: high latitude) stations during storm periods.

\section{Conclusion}
\label{sec:conc}

In this paper, we proposed a scalable inference algorithm for fitting sparse GPR models
with contaminated normal noise. In \cref{sec:parameter-estimation}, we showed that the
algorithm is able to correctly estimate the model hyperparameters when used to fit data
simulated from the model itself. In \cref{sec:comp-to-sparse}, we compared GPR models
with different likelihoods and showed that our proposed model outperforms the others
when there is a large proportion of outliers. In \cref{sec:applications}, we applied our
proposed algorithm to two real-world datasets and discussed strengths and weaknesses
compared to GPR models with other robust likelihoods. 

For the application area of geomagnetic perturbation forecasting, our improvements are two-fold. First, our model was able
to provide more informative prediction intervals for the ground magnetic perturbations dataset when 
compared against other robust methods, including MC Dropout and an ANN with the target objective
of estimating the parameters of a normal distribution. Next, our model consistently avoids 
large over-predictions, especially during periods of high geomagnetic activity. The implementation
of a mixture-based likelihood for the Gaussian Process allows more stable estimates that are
less influenced by outliers at comparable accuracy to state-of-the-art methods.

\section*{Acknowledgements}

This work was supported by the NSF SWQU project PHY 2027555.  YC is further supported by NSF DMS 2113397, NASA 22-SWXC22\_2-0005, and 22-SWXC22\_2-0015. 

\begin{appendix}

\section{Variational inference}
\label{sec:vari-infer}

Variational inference (VI) is commonly used to approximate posterior distributions in
complex Bayesian models where the posterior is not tractable \citep{blei2017variational}.
It provides an alternative to Markov chain Monte-Carlo (MCMC) sampling that
tends to be more computationally efficient and scalable. VI approximates the posterior
distribution by finding the distribution belonging to a simpler family of distributions
that minimizes the Kullback-Leibler (KL) divergence with the exact posterior \citep{blei2017variational}.
More concretely, suppose we have observed data, denoted by $\mathbf{y} \in \mathbb{R}^{n}$, and latent
variables, denoted by $\mathbf{f} \in \mathbb{R}^{m}$. For simplicity, we assume that
$\mathbf{y}$ and $\mathbf{f}$ are real-valued but that is not a requirement for VI. In the
context of GPR, $\mathbf{f}$ are the latent function values.
The goal of VI is to solve the following optimization problem:
\begin{equation}
  q^{*}(\mathbf{f}) = \underset{q(\mathbf{f}) \in \mathcal{Q}}{\text{arg min
    }} \text{KL} \big[q(\mathbf{f}) || p(\mathbf{f} | \mathbf{y}) \big],
  \label{eq:vi-opt-problem}
\end{equation}
where $\mathcal{Q}$ is a family of candidate approximate distributions; and
$p(\mathbf{f} | \mathbf{y})$ is the exact posterior distribution. The KL divergence
between two distributions $q(\mathbf{x})$ and $p(\mathbf{x})$ is defined as
\begin{equation*}
  \text{KL} \big[ q(\mathbf{x}) || p(\mathbf{x}) \big] = \int q(\mathbf{x}) \log \frac{q(\mathbf{x})}{p(\mathbf{x})} d\mathbf{x}.
\end{equation*}
The resulting distribution
$q^{*}(\mathbf{f})$ is then used as an approximation to the exact posterior.
The difficulty of this
optimization problem is dictated by the variational family $\mathcal{Q}$.
Common choices include the mean-field and Gaussian variational family \citep{wainwright2007graphical,opper2009variational}.
The mean-field variational family consists of distributions that asssumes the latent
variables are mutually independent, i.e.
\begin{equation}
  q(\mathbf{f}) = \prod_{j=1}^{m} q_{j}(f_{m}).
  \label{eq:mean-field}
\end{equation}
The Gaussian variational family consists of multivariate Gaussian distributions
parameterized by a mean and covariance matrix. Under this family, the optimization problem
in \cref{eq:vi-opt-problem} simplifies to computing the mean and covariance matrix that
minimizes the KL divergence. In this paper, we will focus on the Gaussian variational
family. The KL divergence in \cref{eq:vi-opt-problem} can be decomposed as
\begin{equation}
  \text{KL} \big[q(\mathbf{f}) || p(\mathbf{f} | \mathbf{y}) \big] = E_{q(\mathbf{f})} [\log q(\mathbf{f})] - E_{q(\mathbf{f})}[\log p(\mathbf{f}, \mathbf{y})] + \log p(\mathbf{y}).
  \label{eq:kl-decomposed}
\end{equation}
The KL divergence depends on the marginal log-likelihood $\log p(\mathbf{y})$ which is
almost always intractable.
We can rearrange \cref{eq:kl-decomposed} to express the marginal log-likelihood as
\begin{equation*}
  \log p(\mathbf{y}) = \text{ELBO}(q) + \text{KL}\big[ q(\mathbf{f}) || p(\mathbf{f} | \mathbf{y}) \big]
\end{equation*}
where
\begin{equation}
  \text{ELBO}(q) = E_{q(\mathbf{f})}[\log p(\mathbf{f}, \mathbf{y})] - E_{q(\mathbf{f})}[\log q(\mathbf{f})].
  \label{eq:elbo-def}
\end{equation}
This function is known as the evidence lower bound (ELBO) because it lower bounds the
marginal log-likelihood (or evidence). This follows from the KL divergence term being
non-negative. Furthermore, since $\log p(\mathbf{y})$ is a constant with respect to $q$,
maximizing the ELBO is equivalent to minimizing the KL divergence between the
approximate and exact posterior, thereby solving the original optimization problem in \cref{eq:vi-opt-problem}.

\section{ELBO for SVGP with Gaussian likelihood without an explicit variational distribution}
\label{sec:elbo-deriv1}

We can rewrite the ELBO derived in \cref{eq:elbo-svgp} as
\begin{equation}
    \text{ELBO} = \int q(\textbf{u}) \log \frac{G(\textbf{u}, \textbf{y}) p(\textbf{u})}{q(\textbf{u})} d\textbf{u},  \label{eq:elbo-int}
\end{equation}
where $\log G(\textbf{u}, \textbf{y}) = \int p(\textbf{f} | \textbf{u}) \log p(\textbf{y} | \textbf{f}) d\textbf{f}$.
If we assume $p(\textbf{y} | \textbf{f}) = N(\textbf{y} | \textbf{f}, \sigma^2 \textbf{I})$ (i.e.\ Gaussian likelihood with homoskedastic noise), then
\begin{align*}
    \log G(\textbf{u}, \textbf{y}) &= \int N(\textbf{f} | \textbf{K}_{nm} \textbf{K}_{mm}^{-1} \textbf{u}, \textbf{K}_{nn} - \textbf{Q}_{nn}) \log [N(\textbf{y} | \textbf{f}, \sigma^{2} \textbf{I})] d\textbf{f} \\
    &= \log[N(\textbf{y} | \textbf{K}_{nm} \textbf{K}_{mm}^{-1} \textbf{u}, \sigma^{2} \textbf{I})] - \frac{1}{2 \sigma^2} \text{Tr}(\tilde{\textbf{K}}_{nn}),
\end{align*}
where $\tilde{\textbf{K}}_{nn} = \textbf{K}_{nn} - \textbf{Q}_{nn}$; and $\text{Tr}(\textbf{A})$ denotes the trace of matrix $\textbf{A}$. Using Jensen's inequality, we can upper bound the ELBO in \cref{eq:elbo-int} with
\begin{equation*}
    \text{ELBO} \leq \log \int G(\textbf{u}, \textbf{y}) p(\textbf{u}) d\textbf{u} = \log N(\textbf{y} | 0, \sigma^2 \textbf{I} + \textbf{Q}_{nn}) - \frac{1}{2 \sigma^2} \text{Tr}(\tilde{\textbf{K}}_{nn}).
\end{equation*}
Therefore, under a Gaussian likelihood, the ELBO for SVGP can be maximized without explicitly assuming a variational distribuiton $q(\textbf{u})$ for the inducing points.

\section{ANN Fitting}
\label{sec:annfit}

Due to computational constraints, hyperparameter tuning for the ANN model is conducted on the OTT station. Hyperparameter tuning station by station for the ANN model would require moving the neural net fitting code to a GPU, a path that was not explored for this paper. Therefore, we estimate that further enhancements to the ANN model's performance are possible. In particular, the hyperparameter optimization performed for the OTT station may not generalize well to other stations, particularly those at higher latitudes. Nonetheless, the current model serves as an adequate baseline, demonstrating comparable performance with prior works such as \cite{keesee2020comparison} and \cite{pinto2022revisiting} when comparing RMSE and skill scores across all test stations.

We opt for a 4-layer deep neural network architecture. Following common convention, we incorporate the rectified linear unit (ReLU) activation function and a batch size of 32. In multivariate regression problems with time series data, it is common to face the issue of gradient explosion. To mitigate this, we utilize the AdamW optimizer with a weight decay of 0.01 and implement gradient clipping with a maximum norm of 1.0. To further reduce overfitting, early stopping is employed after 5 epochs without improvement. The following additional hyperparameters are tuned simultaneously by the Bayesian Optimization library \cite{bayesopt}, by randomly selecting 20 initial combinations of hyperparameters and conducting 40 further iterations. This procedure is repeated multiple times to identify a stable set of hyperparameters. The learning rate is adjusted using an exponential decay schedule, and dropout layers are introduced following each hidden layer at the same rate. The weight of the loss function is also tuned in this process, so we utilize RMSE as the evaluation metric on our validation set. Hyperparameter tuning results are detailed in \cref{tbl:annhyper}.

\begin{table}[ht]
  \centering
  \begin{tabular}{ccccc}
    \toprule
    Layer width & Learning rate & Learning rate decay & Dropout rate & $\lambda$ \\
    \midrule
    100-100-80-80-2 & 0.00008 & 0.8 & 0.1 & 50 \\
    \bottomrule
  \end{tabular}
  \caption{ANN model hyperparameters tuned by Bayesian-Optimization.}
  \label{tbl:annhyper}
\end{table}

Our final model structure and performance results are similar to previous literature of the field \citep{keesee2020comparison,pinto2022revisiting}. However, it is noteworthy that our ANN models tend to produce overly broad confidence intervals, particularly for extreme events and during storm periods. To address this issue, we predict a transformed scale parameter \(\tilde{\sigma}^2\), with the final scale parameter \(\hat{\sigma}^2\) calculated as
\begin{equation*}
\hat{\sigma}^2 = \frac{C}{1 + e^{-A(\tilde{\sigma}^2 - B)}},
\end{equation*}
where \(A\), \(B\), and \(C\) are tunable hyperparameters ensuring that \(\hat{\sigma}^2\) falls within the range \([0,C]\). After exploratory adjustments, these parameters are set to \(A = 0.01\), \(B = 500\), \(C = 1000\) for lower latitude stations (FRD, FRN, NEW, OTT, WNG), and \(C = 2000\) for higher latitude stations (ABK, HRN, IQA, MEA, PBQ, YKC). Though this change reduces interval length in extreme cases, it does not 
eliminate the problem of exploding interval lengths completely.

\begin{table}[h]
\caption{Performance of an ANN model with MC-Dropout. All the parameters are unchanged, except for dropout rate, which is increased to 0.7 to improve its inference capability. 1000 samples are used to estimate the distribution of the final prediction.}
\label{tab:ANN_MC-Dropout}
\centering
\begin{tabular}{lcccc}
\toprule
Station & RMSE & Coverage & Interval mean & Interval median \\
\midrule
NEW & 16.786 & 0.742 & 23.435 & 16.043 \\
OTT & 16.006 & 0.747 & 24.291 & 16.551 \\
WNG & 14.398 & 0.748 & 22.332 & 15.693 \\
YKC & 96.641 & 0.621 & 86.609 & 65.661 \\
ABK & 89.685 & 0.618 & 72.297 & 43.356 \\
MEA & 78.873 & 0.646 & 41.405 & 24.306 \\
FRN & 14.138 & 0.640 & 20.136 & 14.555 \\
IQA & 82.633 & 0.583 & 75.083 & 61.914 \\
PBQ & 87.875 & 0.587 & 75.239 & 48.566 \\
FRD & 11.863 & 0.705 & 19.554 & 14.116 \\
HRN & 65.193 & 0.613 & 70.411 & 56.783 \\
FUR & 12.736 & 0.735 & 19.330 & 13.353 \\
\bottomrule
\end{tabular}
\end{table}

\begin{table}[H]
\centering
\caption{Bias and variance of GPR-CN and ANN models for the testing period of 2015-2016. The ANN method predictions consistently have lower bias and lower variance, at the cost of prediction interval length.}
\label{tbl:bias_var_tradeoff}
\begin{tabular}{lllllllll}
\toprule
 & \multicolumn{4}{l}{Full Set} & \multicolumn{4}{l}{Storm Only} \\
 & \multicolumn{2}{l}{Bias} & \multicolumn{2}{l}{Variance} & \multicolumn{2}{l}{Bias} & \multicolumn{2}{l}{Variance} \\
Model & ANN & GPR-CN & ANN & GPR-CN & ANN & GPR-CN & ANN & GPR-CN \\
Station &  &  &  &  &  &  &  &  \\
\midrule
ABK & 6.061 & 28.750 & 7685.095 & 6578.867 & -2.257 & 43.218 & 21056.316 & 12067.962 \\
FRD & 2.830 & 4.467 & 132.663 & 150.201 & 12.588 & 16.637 & 473.219 & 726.362 \\
FRN & 3.851 & 7.558 & 181.500 & 184.933 & 8.518 & 23.736 & 1065.965 & 793.572 \\
HRN & 12.330 & 19.686 & 4094.969 & 3930.555 & 45.736 & 49.804 & 8905.957 & 8757.528 \\
IQA & -1.114 & 25.646 & 6605.617 & 6383.506 & 15.442 & 52.333 & 11859.860 & 13883.121 \\
MEA & 10.806 & 20.898 & 5275.414 & 5763.874 & 21.143 & 42.103 & 12670.594 & 15761.465 \\
NEW & 3.277 & 6.254 & 258.872 & 309.885 & 10.275 & 15.154 & 1012.890 & 1494.153 \\
OTT & 2.492 & 4.901 & 234.728 & 317.216 & 10.450 & 17.275 & 1553.697 & 2806.177 \\
PBQ & 12.647 & 34.225 & 7184.451 & 7336.389 & 25.398 & 49.459 & 12220.211 & 14443.320 \\
WNG & 2.288 & 4.715 & 188.606 & 202.985 & 9.351 & 15.978 & 787.910 & 971.462 \\
YKC & 5.450 & 26.849 & 8761.938 & 8617.623 & 16.761 & 45.561 & 20696.077 & 23118.061 \\
\bottomrule
\end{tabular}
\end{table}






\end{appendix}











\bibliographystyle{apalike}
\setcitestyle{authoryear}

\bibliography{references.bib}

\end{document}